 \documentclass[pmlr,twocolumn,10pt]{jmlr} 

\usepackage{booktabs}
\usepackage{algorithmic}
\usepackage{siunitx}

\usepackage[switch]{lineno}

\theorembodyfont{\upshape}
\theoremheaderfont{\scshape}
\theorempostheader{:}
\theoremsep{\newline}

\jmlrvolume{297}
\jmlryear{2025}
\jmlrworkshop{Machine Learning for Health (ML4H) 2025} 

 \title[Joint Progression Modeling for Mixed Pathologies]{Joint Progression Modeling (JPM): A Probabilistic Framework for Mixed-Pathology Progression}

  \author{%
\Name{Hongtao Hao}\Email{hhao9@wisc.edu}\\
\addr University of Wisconsin–Madison, USA
\AND
\Name{Joseph L. Austerweil }\Email{joseph.austerweil@gmail.com}\\
\addr \addr Chiba Institute of Technology, Japan \& University of Wisconsin-Madison, USA
\AND
\Name{for the Alzheimer’s Disease Neuroimaging Initiative\textsuperscript{*}}
}

\begin{document}

\maketitle

\begingroup
\renewcommand\thefootnote{*}
\footnotetext{
Data used in preparation of this article were obtained from the Alzheimer’s Disease Neuroimaging Initiative (ADNI) database (\url{adni.loni.usc.edu}). As such, the investigators within the ADNI contributed to the design and implementation of ADNI and/or provided data but did not participate in analysis or writing of this report. A complete listing of ADNI
investigators can be found in Appendix \ref{sec:adni}.}
\endgroup

\begin{abstract}
Event-based models (EBMs) infer disease progression from cross-sectional data, and standard EBMs assume a single underlying disease per individual. In contrast, mixed pathologies are common in neurodegeneration. We introduce the Joint Progression Model (JPM), a probabilistic framework that treats single-disease trajectories as partial rankings and builds a prior over joint progressions. We study several JPM variants (Pairwise, Bradley--Terry, Plackett--Luce, and Mallows) and analyze three properties: (i) \emph{calibration}---whether lower model energy predicts smaller distance to the ground truth ordering; (ii) \emph{separation}---the degree to which sampled rankings are distinguishable from random permutations; and (iii) \emph{sharpness}---the stability of sampled aggregate rankings. All variants are calibrated, and all achieve near-perfect separation; sharpness varies by variant and is well-predicted by simple features of the input partial rankings (number and length of rankings, \emph{conflict}, and \emph{overlap}). In synthetic experiments, JPM improves ordering accuracy by roughly 21\% over a strong EBM baseline (SA-EBM) that treats the joint disease as a single condition. Finally, using NACC, we find that the Mallows variant of JPM and the baseline model (SA-EBM) have results that are more consistent with prior literature on the possible disease progression of the mixed pathology of AD and VaD.
\end{abstract}


\paragraph*{Data and Code Availability}

We used both synthetic and real-world datasets (NACC and ADNI). NACC data can be requested through \url{https://naccdata.org}, and ADNI via \url{https://adni.loni.usc.edu}. We developed a package for JPM, which can be installed via \texttt{pip install pyjpm}. Package source codes can be found at \url{https://github.com/jpcca/pyjpm}. Reproducible codes for the experiments in this study are available at \url{https://github.com/hongtaoh/jpm}.

\paragraph*{Institutional Review Board (IRB)}
The IRB at University of Wisconsin-Madison has reviewed and approved the research (\#2025-1254).

\section{Introduction}
\label{sec:intro}
\begin{table*}[htbp]
\centering
\caption{Participant biomarker measurements}
\label{tab:participant_biomarkers}
\scriptsize{
\setlength{\tabcolsep}{2pt} 
\resizebox{\textwidth}{!}{%
\begin{tabular}{ccccccccccccccccccc}
\toprule
ID & Impacted & ABETA & ADAS13 & CDRSB & Entorhinal & FAQ & FDG & Fusiform & Hippocampus & LDELTOTAL & MMSE & MOCA & MidTemp & PTAU & RAVLT & TAU & TRABSCOR & WholeBrain \\
\midrule
1 & Yes & 846.90 & 13.47 & 1.64 & 0.00 & -0.17 & 1.13 & 0.01 & 0.01 & 0.83 & 30.90 & 18.41 & 0.01 & 11.53 & 44.87 & 378.36 & 124.99 & 0.74 \\
2 & Yes & 1187.20 & 19.92 & 0.04 & 0.00 & -0.66 & 1.26 & 0.02 & 0.01 & 5.22 & 29.12 & 22.72 & 0.01 & 12.32 & 53.85 & 113.02 & 112.70 & 0.68 \\
3 & No & 923.78 & 13.40 & -0.10 & 0.00 & 2.07 & 1.30 & 0.02 & 0.00 & 7.18 & 30.05 & 22.65 & 0.01 & 21.54 & 43.63 & 228.85 & 38.96 & 0.70 \\
4 & Yes & 762.25 & 7.75 & 0.79 & 0.00 & -0.37 & 1.22 & 0.02 & 0.01 & 2.66 & 28.85 & 19.61 & 0.01 & -3.13 & 21.16 & 541.09 & 78.48 & 0.68 \\
5 & No & 506.22 & 20.85 & 0.24 & 0.00 & 0.48 & 1.37 & 0.01 & 0.00 & 3.98 & 28.66 & 26.52 & 0.01 & 14.58 & 24.91 & 185.07 & 114.84 & 0.68 \\
\bottomrule
\end{tabular}
}}
\par\vspace{0.2cm}
\textit{Note:} Biomarker names are abbreviated for brevity. Values of 0.0 result from rounding to two decimal places. Detailed biomarker descriptions are provided in Table~\ref{tab:biomarker_glossary_extended} (Appendix~\ref{apd:biomarkers}).
\end{table*}

Modern neuroscience often models the progression of neurodegenerative diseases (NDDs) with frameworks like the Event-Based Model \citep[EBM,][]{fonteijn2012event,young2024data}. These models have provided critical insights into the sequential evolution of biomarkers from cross-sectional data~\citep{young2014data, young2018uncovering, archetti2019multi, venkatraghavan2019disease}. They have been successfully applied across multiple diseases, such as Alzheimer's (AD), frontotemporal dementia, Huntington's, and Parkinson's~\citep{young2018uncovering, fonteijn2012event, oxtoby2021sequence, wijeratne2023temporal, firth2020sequences}. However, they are built on a simplifying, yet unrealistic, assumption: Each patient suffers from a single, isolated pathology. This assumption is inconsistent with decades of neuropathological evidence. The clinical reality is that patients often have multiple diseases (mixed pathology, or comorbidity), especially in the case of  NDDs~\citep{rahimi2014prevalence, forrest2023current, buchman2021person, chu2023mixed, jorgensen2020age, stirland2025multimorbidity, jellinger2021pathobiological}. Autopsy studies show that 45.8\% of clinically probable AD patients have mixed pathologies such as infarcts, and Lewy bodies~\citep{schneider2009neuropathology}. The rate of mixed pathologies is even higher in Progressive Supranuclear~\citep[70\%;][]{popli2025high} and Parkinson's~\citep[84\%;][]{buchman2021person}. How can we build models of how diseases jointly progress based on individual disease progression?

To illustrate the challenge of modeling mixed pathology progression, consider Table~\ref{tab:participant_biomarkers}. It contains data of 5 participants (from a larger data set): 3 impacted and 2 healthy. Impacted participants have all triple pathologies: AD, Vascular Dementia (VaD), and Lewy Body Dementia (LBD). Healthy participants are free from all of them. Biomarkers involved in each of the three pathologies are available at Appendix~\ref{apd:biomarkers_disease}, and their measurements are continuous. Now the question: what is the ordinal progression of this comorbidity? We define a disease progression as the ordinal order in which biomarkers get pathological. For example, ABETA $\to$ TAU $\to$ MOCA $\cdots$. Note that we assume all individuals share a common progression for a specific disease. 

A seemingly straightforward solution would be to treat comorbidity as a unique disease and apply existing models such as EBM from scratch. This approach, however, is both data-inefficient and practically infeasible. It discards the wealth of information available in large single-disease cohorts and imposes unrealistic data requirements. The scale of this data-sparsity problem is starkly illustrated by the National Alzheimer’s Coordinating Center (NACC) dataset, one of the world's most comprehensive clinical collections for AD and related dementias. While it contains over 13,400 participants with AD alone, only 50 participants with a mixed diagnosis of AD and VaD, and also have the cerebrospinal fluid (CSF) biomarkers essential for modern AD diagnosis and prognosis. (Participants in studies focusing on VaD do not always collect CSF as it can be costly and painful to do so). It is statistically untenable to build a robust progression model from such a small or incomplete sample. 

In this study, we introduce the Joint Progression Model (JPM), a comprehensive probabilistic framework designed to model mixed pathologies by integrating knowledge from single-disease cohorts. Critically, the JPM is designed for both generative and inferential tasks. This enables not only the generation of synthetic data to probe disease mechanisms and validate models, but also the robust inference of progression and patients' disease stages. The framework's core connects joint progression to the rich theoretical literature on probabilistic ranking, e.g., \cite{mallows1957non} and \cite{plackett1975analysis}, using these established models to build a prior within a flexible Bayesian framework that can be paired with any disease progression likelihood function, such as the EBM. 

In this paper, we make the following contributions. Firstly, we develop the JPM as a theoretical framework for mixed-pathology modeling using the EBM as its likelihood. Secondly, we formally analyze the statistical properties of the JPM, explore how it connects to real-world scenarios, and identify the conditions required for accurate and reliable inference. Lastly, we apply the JPM to both synthetic and real-world datasets, demonstrating up to a 21\% improvement in progression estimation accuracy compared to the state-of-the-art single-disease EBM~\citep{hao2025stage}.

\section{Related work}
The EBM~\citep{fonteijn2012event} was a seminal contribution, enabling the inference of biomarker event sequences from cross-sectional data. A family of variants followed, extending the EBM to handle patient heterogeneity~\citep{venkatraghavan2019disease}, discover disease subtypes~\citep{young2018uncovering,hao2025subtypes}, and scale to high-dimensional data~\citep{tandon2023sebm, wijeratne2024unscrambling}. While alternative frameworks using deep learning on longitudinal data exist for diagnosis predictions~\citep{lee2019predicting, yang2021deep}, they do not recover event sequences. Crucially, all prior work assumes a single pathology. 

This single-disease focus is a major limitation, as co-neuropathologies are highly prevalent and clinically significant~\citep{schneider2009neuropathology, forrest2023current}. To date, modeling mixed pathology progression remains challenging.~\citep{young2024data}. The common practices are to exclude comorbid patients from studies, e.g., in ~\cite{jorgensen2020age}, or predict patient state conversion, e.g., healthy to diseased ~\citep{oflaz2023,maag2021modeling}. Both practices do not allow us to  understand how diseases interact as they progress jointly.

We propose to fill this gap by applying probabilistic ranking theory~\citep{tang2019mallows}. There exist several possible probabilistic models over rankings or orderings, such as the Mallows~\citep{mallows1957non}, Bradley-Terry~\citep{bradley1952rank}, and Plackett-Luce models~\citep{luce1959individual, plackett1975analysis}. While widely used in other fields such as election~\citep{gormley2009grade}, consumer preference~\citep{feng2022mallows}, and information retrieval~\citep{liu2009learning}, these models have not yet been applied to joint disease progression. Below, we demonstrate how we integrate these models into a generative and inference framework to tackle the mixed-pathology problem.

\section{JPM theory and algorithms}

\subsection{Problem statement}

We define a disease progression as the ordinal sequence in which relevant biomarkers become pathological. We call the progression of an individual disease as a \textbf{partial ranking}, and the progression of the comorbidity as an \textbf{aggregate ranking}. Consider two diseases with their respective progressions: 

\begin{itemize}
    \item Disease 1: A$\beta$ $\to$ Tau $\to$ cognitive decline $\to$ hippocampus atrophy
    \item Disease 2: white matter lesions (WML) $\to$ hippocampus atrophy $\to$ cognitive decline
\end{itemize}

The challenge is to determine the most probable aggregate ranking for patients with both diseases.  WML $\to$ A$\beta$ $\to$ Tau $\to$ hippocampus atrophy $\to$ cognitive decline is probably more likely than cognitive decline $\to$ atrophy $\to$ WML $\to$ Tau $\to$ A$\beta$. But by how much? Is it also more probable than WML $\to$ cognitive decline $\to$ A$\beta$ $\to$ Tau $\to$ hippocampus atrophy? A sophisticated model is needed to address these questions when we have multiple partial rankings with biomarkers whose order may conflict across them.

Formally, let $\mathcal{X}^{(k)}$ denote the set of biomarkers associated with disease $k$, with $k \in \{ 1, \ldots, K \}$. Let $\mathcal{S} = \left( \sigma^{(1)}, \sigma^{(2)}, ..., \sigma^{(K)}\right)$ denote the $K$ partial rankings, each a total ordering encoding the corresponding (individual) disease's progression over its biomarkers. Note that this is a partial ranking from the perspective of the comorbidity. Let

$$\mathcal{X} = \bigcup_{k=1}^{K} \mathcal{X}^{(k)}$$

\noindent be the set of all unique biomarkers, with $m=|\mathcal{X}|$. Any aggregate ranking $\sigma$ is a permutation of $\mathcal{X}$. 

JPM aims to obtain the posterior of aggregate rankings ($\sigma$) given the observed mixed-pathology data ($D$) and partial rankings ($\mathcal{S}$):

$$
P(\sigma \mid D, \mathcal{S}) 
\propto P(D \mid \sigma, \mathcal{S}) \, P(\sigma \mid \mathcal{S})$$

\noindent Because $\mathcal{S}$ is fixed ground truth, we can drop it from the data likelihood:

$$P(\sigma\mid D, \mathcal{S}) \propto P(D\mid \sigma) \cdot P(\sigma\mid \mathcal{S})$$

We adopt an energy-based formulation for the prior $P(\sigma\mid\mathcal{S})$:

$$P(\sigma \mid \mathcal{S}) \propto \exp \left(- E(\sigma \mid \mathcal{S}) \right)$$

\noindent where $E(\sigma \mid \mathcal{S})$ is the \textbf{energy function} that assigns a low value to more probable rankings. This exponential form ensures positivity and allows the model to concentrate probability mass around low-energy (i.e., high-likelihood) rankings. Below we describe four principled choices for this energy function

\subsection{Variant 1: Pairwise Preferences (PP)}

The Pairwise Preferences model generalizes majority voting across partial rankings. It treats the energy of aggregate ranking as the outcome of weighted pairwise comparisons between biomarkers. 

$$E(\sigma \mid \mathcal{S}) = - \sum_{i,j \in \mathcal{X}, i \neq j} w_{ij} \mathbf{1}[i <_{\sigma} j]$$

\noindent where 

\begin{itemize}
    \item $\mathbf{1}[i <_{\sigma} j] = 1$ if $i$ precedes $j$ in ranking $\sigma$ (i.e., $i \prec j$); otherwise $0$.
    \item $w_{ij}$ is the weight of preference for $i$ preceding $j$.
\end{itemize}

We derive $w_{ij}$ from partial rankings:

$$w_{ij} = \sum_{k=1}^K \lambda_k \cdot r_k(i, j), \quad \forall i,j \in \mathcal{X}, i \neq j$$

\noindent with: 

\begin{itemize}
    \item $\lambda_k \ge 0$ is the importance of $\sigma^{(k)}$ (default $\lambda_k = 1$).
    \item If both $i$ and $j$ are present in $\sigma^{(k)}$: $r_k (i,j) = 1$ when $i \prec j$, and $r_k (i,j) = -1$ when $j \prec i$.
    \item $r_k (i,j) = 0$ otherwise.
\end{itemize}

\subsection{Variant 2: Generalized Bradley--Terry (BT)}

The Bradley-Terry model~\citep{bradley1952rank} interprets rankings as arising from latent strength parameters assigned to each item. Each item $i \in \mathcal{X}$ is associated with a parameter $\theta_i$, forming a parameter vector $\boldsymbol{\theta}$. The probability of item $i$ preceding $j$ is:

$$
P(i \prec j) = \frac{e^{\theta_i}}{e^{\theta_i} + e^{\theta_j}}
$$

Let $T^{(k)}$ be the set of all unique items in a partial ranking $\sigma^{(k)} = \left(\sigma_1^{(k)}, \sigma_2^{(k)}, \ldots, \sigma_{n_k}^{(k)} \right)$. For any pair $i, j \in T^{(k)}$, let $C_{i,j}^{(k)} = 1$ if $i \prec j$, i.e., $i$ is ranked higher than $j$ in partial ranking $\sigma^{(k)}$; otherwise $C_{i,j}^{(k)} = 0$. The likelihood of observing the ranking $\sigma^{(k)}$ given $\boldsymbol{\theta}$ is:

$$
p(\sigma^{(k)} \mid \boldsymbol{\theta}) = \prod_{\substack{i,j \in T^{(k)} \\ i \prec j}} P(i \prec j)^{C_{i,j}^{(k)}}
$$

Taking the logarithm, we obtain the log-likelihood:

$$
\log p(\sigma^{(k)} \mid \boldsymbol{\theta}) = \sum_{\substack{i,j \in T^{(k)} \\ i \prec j}} C_{i,j}^{(k)} \cdot \left( \theta_i - \log(e^{\theta_i} + e^{\theta_j}) \right)
$$








Maximizing the total log-likelihood yields parameter estimates:

$$\hat{\vec{\theta}} = \arg\max_\theta \sum_{k=1}^K \log P(\sigma^{(k)} | \vec{\theta})$$

The energy for an aggregate ranking $\sigma$ is:

$$
E(\sigma \mid \mathcal{S}) = E(\sigma \mid \hat{\vec{\theta}}) = - \log P(\sigma^{(k)} \mid \vec{\hat{\theta}})
$$

\subsection{Variant 3: Plackett--Luce (PL)}

The Plackett-Luce model~\citep{plackett1975analysis, luce1959individual} generates rankings sequentially, choosing one item at a time with probability proportional to an associated score. Each item $i \in \mathcal{X}$ has a parameter $\alpha_i$. For a partial ranking $\sigma^{(k)} = [x_1, x_2, \ldots, x_{n_k}]$, where $n_k \le m$ and $\{ x_1, x_2, \ldots, x_{n_k}\} \subseteq \mathcal{X}$:

$$
P(\sigma^{(k)} \mid \vec{\alpha}) = \prod_{t=1}^{n_k} \frac{e^{\alpha_{x_t}}}{\sum_{j=t}^{n_k} e^{\alpha_{x_j}}}
$$

The log-likelihood is:

$$
\log P(\sigma^{(k)} \mid \vec{\alpha}) = \sum_{t=1}^{n_k} \left( \alpha_{x_t} - \log \sum_{j=t}^{n_k} e^{\alpha_{x_j}} \right)
$$

Parameters are estimated via maximum likelihood:

$$\hat{\vec{\alpha}} = \arg\max_\alpha \sum_{k=1}^K \log P(\sigma^{(k)} | \vec{\alpha})$$

For an aggregate ranking $\sigma = [y_1, y_2, \ldots, y_m]$, the energy is:

$$
E(\sigma \mid \mathcal{S}) = E(\sigma \mid \vec{\hat{\alpha}}) = - \log P(\sigma^{(k)} \mid \vec{\hat{\alpha}})
$$

\subsection{Variant 4: BT-informed Mallow's Model}
\label{subsec:mallows}

The Mallows model~\citep{mallows1957non} places probability mass around a central ranking $\sigma_0$: 

$$P(\sigma) \propto \exp(-\theta \cdot d(\sigma, \sigma_0))$$ 

\noindent where $d(\cdot,\cdot)$ is a distance metric (e.g., Kendall's tau or Reverse Major Index \citep[RMJ;][]{feng2022mallows}). Dispersion $\theta$ controls how concentrated the distribution is.  The challenge is inferring $\theta$ and $\sigma_0$ from a small number of heterogeneous partial rankings. Existing methods, for instance, BayesMallows~\citep{vitelli2018probabilistic, sorensen2019bayesmallows}, require richer data and are unstable when only 2-4 partial rankings are available---typical in mixed-pathology settings. In our preliminary experiments, estimated central rankings given a fixed set of partial rankings varied greatly across runs (Kendall's W $\approx$ 0.1), while BT produced stable estimates ($>0.9$). We decide to use the inference method from BT to provide an estimate of $\sigma_0$ and set $\theta$ manually, which we call \emph{BT-informed Mallows model}. The energy for the BT-informed Mallows model is: 

$$E(\sigma \mid \mathcal{S}) = \theta \cdot d(\sigma, \sigma_0)$$ 

\noindent where the dispersion $\theta$ provides a flexible knob for synthetic data generation: larger $\theta$ yields a more peaked distribution on the central ordering $\sigma_0$, and smaller $\theta$ yields a more uniform distribution over $\sigma$.

\subsection{Generative and inference algorithms}
\label{sec:generative_and_inference_algo}

JPM can be used for both generative and inferential purposes. With a defined JPM, we can define algorithms for generating aggregate rankings based on input partial rankings, which is crucial for creating synthetic data. We can also infer the aggregate rankings from a data set (Algorithm~\ref{alg:jpm_algo} in Appendix~\ref{apd:sampling}).  In terms of generating aggregate rankings, PL sampling is straightforward, as rankings are generated sequentially (See Algorithm~\ref{alg:pl_sample}). Other models (PP, BT, Mallows) do not admit direct sampling. We employ Metropolis-Hastings MCMC~\citep{metropolis1953equation, hastings1970monte}, starting from a random permutation and proposing swaps between positions (See Algorithm~\ref{alg:mh_sample}). Note that MCMC can be applied to PL as well. Throughout the paper, we used Kendall's $\tau$ as the distance metric.

\section{Theoretical analysis}
We explored five key questions regarding the JPM framework:
\begin{itemize}
    \item[\textbf{Q1:}] Is JPM valid as an inference model?
    \item[\textbf{Q2:}] Is JPM valid as a generative model?
    \item[\textbf{Q3:}] Which JPM variants are suitable for synthetic data generation?
    \item[\textbf{Q4:}] For a given dataset, should JPM be preferred over a single-disease model for inference (e.g., SA-EBM~\citep{hao2025stage})?
    \item[\textbf{Q5:}] If JPM is appropriate for inference, which variant should be used?
\end{itemize}

\noindent This section addresses the first three questions. Q4 and Q5 will be addressed in the Discussion.

\subsection{Q1: Model calibration}

JPM as an inference algorithm aims to correctly assign energies to different aggregate rankings. We define a ``valid inference model'' as one that assigns low energies to aggregate rankings with smaller distances from the ground truth aggregate ranking ($\sigma_{\text{gt}}$). Formally, for any two candidate rankings $\sigma_A$ and $\sigma_B$, a well-behaved model should satisfy the condition:
$$E_{\text{JPM}}(\sigma_A) < E_{\text{JPM}} (\sigma_B) \Longrightarrow d(\sigma_A, \sigma_{\text{gt}}) < d(\sigma_B, \sigma_{\text{gt}})$$

We refer to the degree to which a model satisfies this property as its \textbf{calibration}. We used the 4,050 aggregate rankings generated in Sec.~\ref{sec:synthetic_data_generation} to study model calibration. We treat each aggregate ranking as an experiment. In each experiment, the given aggregate ranking served as the ground truth ($\sigma_{\text{gt}}$), and we also generated 1,000 random aggregate rankings based on the original associated input partial rankings which generated this $\sigma_{\text{gt}}$. As discussed in Sec.~\ref{sec:generative_and_inference_algo}, we can use different JPM variants to generate an aggregate ranking from a set of partial rankings. Therefore, in each experiment, we tried each of the four inference variants: BT, PL, PP and Mallows ($\theta = 1$) when generating the 1,000 random aggregate rankings.

We measured \emph{calibration} by calculating the Spearman's $\rho$ correlation between (1) the energies of 1,000 randomly generated aggregate rankings, and (2) the average Kendall's $\tau$ distance between those random aggregate rankings and $\sigma_{\text{gt}}$. 


\begin{table}[htbp]
\centering
\caption{Calibration of JPM variants (mean with 95\% CI). See Fig.~\ref{fig:clb} in Appendix~\ref{apd:model_analysis} for details.}
\label{tab:clb}
\resizebox{\linewidth}{!}{%
\begin{tabular}{lcccc}
\toprule
\textbf{Generative variant} & \textbf{BT} & \textbf{Mallows ($\theta=1$)} & \textbf{PL} & \textbf{PP} \\
\midrule
BT              & 0.802 $\pm$ 0.009 & 0.977 $\pm$ 0.001 & 0.605 $\pm$ 0.007 & 0.884 $\pm$ 0.003 \\
Mallows ($\theta=1$)  & 0.783 $\pm$ 0.010 & 0.954 $\pm$ 0.002 & 0.590 $\pm$ 0.007 & 0.866 $\pm$ 0.003 \\
Mallows ($\theta=10$) & 0.780 $\pm$ 0.010 & 0.950 $\pm$ 0.002 & 0.579 $\pm$ 0.007 & 0.862 $\pm$ 0.003 \\
PL              & 0.849 $\pm$ 0.007 & 0.913 $\pm$ 0.004 & 0.660 $\pm$ 0.006 & 0.840 $\pm$ 0.005 \\
PP        & 0.791 $\pm$ 0.009 & 0.944 $\pm$ 0.002 & 0.590 $\pm$ 0.007 & 0.913 $\pm$ 0.002 \\
\bottomrule
\end{tabular}%
}
\end{table}

As shown in Table~\ref{tab:clb}, all JPM variants achieve a \emph{calibration} greater than 0.58, indicating a positive correlation. The Mallows variant scores the highest, which is expected as its energy function is explicitly defined by Kendall's $\tau$. \textbf{In sum, Table~\ref{tab:clb} indicates that JPM variants may all be valid inference algorithms}, because they can theoretically identify rankings closer to the ground truth. 

While the \emph{calibration} results in Table~\ref{tab:clb} are promising, JPM does not guarantee the accuracy of the final inferred result, as the search process during inference may not explore the entire space of possible rankings. Consider this concrete example: JPM explores permutations that are all far from the ground truth. However, JPM correctly finds that permutations that are further away from the ground truth have larger distances, i.e., good \emph{calibration} scores. This way, valid inference algorithms could be created from a calibrated JPM, but a specific inference algorithm may still fail to find good results.

\subsection{Q2: Characterizing the Generation of Aggregate Rankings}

In Q1 experiments, we generated aggregate rankings based on partial rankings, using JPM as a generative framework. We can conclude that in the real-world, if the generative process can be accurately represented by a JPM, then JPM may be a valid inference model (because of the good \emph{calibration} scores). This leads to the next question: \textbf{How well can JPM's generative process emulates the formation of real-world joint disease progressions?} We analyze this from two perspectives: \textbf{separation} and \textbf{sharpness}.

\emph{Separation} assesses whether a model's energy landscape clearly distinguishes the true aggregate ranking ($\sigma_{\text{gt}}$) from random noise. In other words, the ground truth aggregate ranking must be more consistent with the evidence from the input partial rankings ($\mathcal{S}$) than a random ranking is. Formally:
$$E(\sigma_{\text{gt}}\mid \mathcal{S}) \ll E(\sigma_{\text{random}}\mid \mathcal{S})$$

We define \emph{separation} using the Area Under the Receiver Operating Characteristic Curve (AUROC), which measures the probability that a ranking sampled from the JPM has a lower energy than a randomly generated one:

$$\text{separation} = P(E_{\text{JPM}} < E_{\text{rand}}) + \frac{1}{2} P(E_{\text{JPM}} = E_{\text{rand}}),$$

\noindent where $E_{\text{JPM}}$ represents the energies of rankings sampled by JPM, and $E_{\text{rand}}$ represents the energies of random permutations of the biomarkers in $\mathcal{X}$.

\emph{Sharpness} addresses a related question: Given a set of input partial rankings ($\mathcal{S}$), how stable is the generative output? A model with high \emph{sharpness} will consistently produce similar aggregate rankings across repeated runs. We define \emph{sharpness} as the Kendall's $W$ of the sampled aggregate rankings.

We analyzed \emph{separation} and \emph{sharpness} also using the 4,050 aggregate rankings generated in Sec.~\ref{sec:synthetic_data_generation}, but in a different way than we did in the study of \emph{calibration}. To study \emph{separation}, for each of the five generative variants (as in Table~\ref{tab:sep_shp}), in each of the 4,050 ``experiments'', we generated 1,000 aggregate rankings using the variant, and 1,000 random permutations of the biomarkers in $\mathcal{X}$. 

As shown in Table~\ref{tab:sep_shp}, all JPM models exhibit high \emph{separation}, but they differ significantly in their \emph{sharpness}. Notably, the Mallows variant's \emph{sharpness} approaches 1.0 as its parameter $\theta$ increases. This result carries two important implications. First, JPM is best suited for generative tasks where the underlying progression is meaningfully different from a random ordering (i.e., high \emph{separation}). Second, assuming the high-separation condition is met, the JPM framework offers the flexibility to generate synthetic data with a wide range of \emph{sharpness}, allowing it to simulate diverse real-world scenarios.

\begin{table}[htbp]
\centering
\caption{Mean separation and sharpness by data generation framework (mean $\pm$ 95\% CI). See Fig.~\ref{fig:sep} in Appendix~\ref{apd:model_analysis} for details.}
\label{tab:sep_shp}
\resizebox{\linewidth}{!}{%
\begin{tabular}{lcc}
\toprule
\textbf{Generative variant} & \textbf{Separation} & \textbf{Sharpness} \\
\midrule
BT              & 1.00 $\pm$ 0.00 & 0.96 $\pm$ 0.00 \\
Mallows ($\theta=1$)  & 1.00 $\pm$ 0.00 & 0.47 $\pm$ 0.00 \\
Mallows ($\theta=10$) & 1.00 $\pm$ 0.00 & 0.61 $\pm$ 0.00 \\
PL              & 1.00 $\pm$ 0.00 & 0.74 $\pm$ 0.01 \\
Pairwise        & 1.00 $\pm$ 0.00 & 0.86 $\pm$ 0.00 \\
\bottomrule
\end{tabular}%
}
\end{table}


\subsection{Q3: Which Generative Variant to Choose?}
\label{sec:q3}

Our analysis of the simulation data reveals that for every variant except the PP model, the \emph{sharpness} of the generated output is highly predictable from the characteristics of the input partial rankings (see Table~\ref{tab:combined_sharpness} in Appendix~\ref{apd:model_analysis}). We identify four key predictive characteristics: the number of partial rankings ($K$), the average length of the partial rankings ($\bar{\ell}$), and two measures we call \textbf{conflict} and \textbf{overlap}.

We define \emph{conflict} as the average pairwise normalized Kendall's $\tau$ distance (See Algorithm~\ref{alg:normalized_tau_distance} in Appendix~\ref{apd:normalized_tau_distance}) among the input partial rankings, restricted to their common items:
\[
\text{Conflict} = \frac{2}{K(K-1)} \sum_{1 \leq i < j \leq K} d_\tau\!\bigl(\sigma^{(i)}, \sigma^{(j)}\bigr)
\]

We define \emph{overlap} as the proportion of biomarkers that appear in at least two partial rankings relative to the total number of unique biomarkers, $m$:

\[
\text{Overlap}
= \frac{
\left|\displaystyle \bigcup_{1 \le i < j \le K}
\left(\sigma^{(i)} \cap \sigma^{(j)}\right)\right|
}{m}.
\]

This provides a clear guide for model selection. A user can first determine a target \emph{sharpness} that suits their research needs. Then, by calculating these four characteristics from their input data and plugging them into regression models (as illustrated in Table~\ref{tab:combined_sharpness}), they can select the JPM variant that will produce an output with the desired sharpness. For a detailed breakdown of how each characteristic influences \emph{sharpness}, see Figures~\ref{fig:bt}, \ref{fig:pl}, \ref{fig:pp}, \ref{fig:m1} and \ref{fig:m10} in Appendix~\ref{apd:model_analysis}.

\section{Synthetic experiments \& results}
To evaluate the JPM framework, we designed a series of experiments using synthetic data. The pipeline for these experiments involved two main stages: (1) generating realistic synthetic datasets under various conditions, and (2) applying and evaluating both the JPM and the SA-EBM \citep[for comparison,][]{hao2025stage} on these datasets.

\subsection{Synthetic data generation}
\label{sec:synthetic_data_generation}

Our synthetic data generation process was designed to simulate plausible clinical research scenarios. We selected 18 biomarkers from Alzheimer's Disease Neuroimaging Initiative \citep[ADNI,][]{mueller2005alzheimer}: 12 are commonly used in AD progression modeling studies \citep{young2014data, archetti2019multi}, and an additional 6 to ensure pathological diversity. The ADNI was launched in 2003 as a public-private partnership, led by Principal Investigator Michael W. Weiner, MD. The primary goal of ADNI has been to test whether serial magnetic resonance imaging (MRI), positron emission tomography (PET), other biological markers, and clinical and neuropsychological assessment can be combined to measure the progression of mild cognitive impairment (MCI) and early Alzheimer’s disease (AD). In Appendix~\ref{apd:biomarkers}, we detailed the list of these 18 biomarkers and how we obtained their distribution parameters. 

To generate one synthetic dataset, we first generated 2-4 randomized partial rankings, each with 6-12 biomarkers randomly chosen from the 18 biomarkers mentioned above. These ranges were chosen to reflect the practical constraints observed in the real-world NACC data. Based on this set of partial rankings, we used JPM variants (BT, PL, PP, and Mallows with $\theta=1,10$) to generate a ground truth aggregate ranking. If MCMC was used (Algorithm~\ref{alg:mh_sample}), we ran it for 500 iterations. We then generated the final biomarker data using two different generative models: (1) Event-Based Model \citep[EBM;][]{fonteijn2012event}; (2) Sigmoid Model (See details in Appendix~\ref{apd:sigmoid}), which was inspired by a hypothetical AD cascade model~\citep{jack2010hypothetical} and adapted from prior work \citep{venkatraghavan2019disease, young2015simulation}.

We varied the following in data generation. (1) Participant size ($J$): 50, 100, or 200 participants. (2) Healthy ratio ($R$), i.e., the percent of control participants: 0.25, 0.5, or 0.75. (3) Experimental configurations (See Appendix~\ref{apd:experimental_setup} for the full list). These include variations in disease stage type (ordinal vs. continuous) and distributions (e.g., Dirichlet-Multinomial vs. Uniform), and biomarker data distributions (Normal vs. Non-Normal).

For each of the $3 \times 3 = 9$ combinations of participant size ($J$) and healthy ratio ($R$), we generated 10 random datasets. In total, considering 5 JPM variants (Mallows with two temperature settings) and 9 experimental conditions, this resulted in $5 \times 9 \times 90 = 4050$ datasets for the aggregate disease scenarios. In parallel, we generated corresponding datasets for each individual disease. The configuration for these datasets matched the aggregate one, but the participant size was set to be \textbf{four times larger}, a design choice also inspired by properties of the NACC data.

\subsection{Experiment setup and evaluation}

For each synthetic aggregate dataset, we first ran SA-EBM~\citep{hao2025stage} on the corresponding individual disease datasets to estimate their respective partial rankings. These estimated rankings were then used to construct a JPM, whose result served as an informed prior (as in Algorithm~\ref{alg:jpm_algo}) for a final SA-EBM run on the aggregate, i.e., mixed-pathology dataset. As a benchmark, we ran SA-EBM alone without the JPM-resultant prior. We chose SA-EBM because it has the state-of-the-art performance on reconstructing ordinal disease progressions from cross-sectional datasets~\citep{hao2025stage}. For all SA-EBM and JPM inference methods, we used 20,000 iterations for MCMC.

We evaluated performance on two tasks. (1) Ordering: The agreement between the estimated ranking and the ground truth, measured by the normalized Kendall's $\tau$ distance (See Algorithm~\ref{alg:normalized_tau_distance} in Appendix~\ref{apd:normalized_tau_distance}). A lower value indicates better performance; (2) Staging: The accuracy of participant staging, measured by the Mean Absolute Error (MAE).

\subsection{Results}

The results are grouped by the model used to generate the synthetic data. If the Mallows model variant was used for inference, i.e., reconstructing the mixed-pathology progression and assigning participants disease stages, we used a fixed $\theta=1$.

\paragraph{Data Generated by PP}
When data was generated using the PP model (See Figures~\ref{fig:pp_tau} and \ref{fig:pp_mae} in Appendix~\ref{apd:synthetic_results}), JPM demonstrated a clear and significant advantage over the single-disease EBM. Error is reported as 95\% confidence intervals throughout the paper. The JPM-EBM achieved the best ordering performance with Kendall's $\tau$ of \textbf{$0.15 \pm 0.01$}, outperforming SA-EBM ($0.19 \pm 0.01$). The PL and BT priors also performed well ($0.17 \pm 0.01$). This advantage was most pronounced for smaller sample sizes ($J=50, 100$), where a prior is most impactful. In the staging task, JPM also outperformed SA-EBM ($0.79\pm 0.07$ MAE), with the PP achieving the best result ($0.75 \pm 0.07$ MAE).

\paragraph{Data Generated by BT}
JPM also showed a significant advantage when the data were generated using the BT model. The BT, PL, and PP variants achieved the best ordering results (\textbf{$0.15 \pm 0.01$}). The Mallows model performed comparably to SA-EBM ($0.19 \pm 0.01$). Staging results were similar to those from the PP-generated data, with BT having the best performance. See Figures~\ref{fig:bt_tau} and \ref{fig:bt_mae} in Appendix~\ref{apd:synthetic_results}.

\paragraph{Data Generated by PL}
When data were generated with the PL model, the performance advantage of JPM on the ordering task was smaller but still present. The PL variant achieved the best outcome ($\tau = 0.17 \pm 0.01$), slightly better than the standard EBM ($0.20 \pm 0.01$). The margin of advantage in the staging task was similar to the above two experiments ($0.07$ MAE). See Figures~\ref{fig:pl_tau} and \ref{fig:pl_mae} in Appendix~\ref{apd:synthetic_results}.

\paragraph{Data Generated by the Mallows Variant}
When data were generated using Mallows ($\theta=1.0$), on the ordering task, the Mallows variant performed similarly to SA-EBM ($0.19 \pm 0.01$), only slightly better than PL ($0.20 \pm 0.01$) and PP ($0.21 \pm 0.01$). The BT variant performed the worst in this condition ($0.23 \pm 0.01$). A similar pattern was observed when data was generated with a higher concentration ($\theta=10.0$). Staging results on Mallows-generated data reflected the patterns seen in the ordering task, i.e., SA-EBM and the Mallows model were the top performers. See Figures~\ref{fig:m1_tau}, \ref{fig:m1_mae}, \ref{fig:m10_tau} and \ref{fig:m10_mae} in Appendix~\ref{apd:synthetic_results} for more details.

\section{NACC experiment \& results}
We obtained data from the National Alzheimer’s Coordinating Center (NACC) and the Alzheimer’s Disease Sequencing Project (ADSP)~\citep{mukherjee2023cognitive}. We used Uniform Data Set (UDS) visits conducted between September 2005 and June 2025. In Appendix~\ref{apd:nacc_data}, we detailed what data were used, how we processed the data, and the information about biomarkers that we included for AD, VaD and the mixed-pathology of AD and VaD. 

The resulting dataset of AD contains 188 healthy participants and 59 AD patients. VaD dataset contains 3,732 healthy participants and 525 VaD patients. The mixed-pathology dataset contains 188 healthy participants and 37 patients with both AD and VaD. 

To get the progression of AD, we ran SA-EBM~\citep{hao2025stage} ten times; each time with a random seed. We obtained the final result with the seed associated with the highest log data likelihood. The VaD dataset is too large to run with multiple seeds. We ran with two seeds and obtained the result which seems more plausible. For mixed-pathology, we first ran with SA-EBM with ten random seeds and obtained the result using the highest-likelihood seed. We also tried four variants of the JPM. For the Mallows model, we used $\theta=1$. For each variant, we also tried 10 random seeds and obtained the result using the seed associated with the highest data log likelihood. In all runs, we used 20,000 MCMC iterations with 500 burn in and no thinning. Detailed results, including the progressions and the trace plots, are presented in Appendix~\ref{apd:nacc_results}.

The timelines of the baseline SA-EBM model and the JPM-Mallows model align better with the current understanding of how vascular disease interacts with Alzheimer's pathology. They positioned vascular damage (WMHnorm) as an event that occurred after the initial molecular pathology was present and cognitive symptoms were just beginning, but before the cascade of major, widespread brain volume loss. This sequence supports a synergistic role for vascular disease, where it exacerbates the ongoing Alzheimer's pathology, accelerating the subsequent neurodegeneration and brain atrophy~\citep{attems2014overlap, snyder2015vascular, lee2016white}. Between the two results, the one by JPM-Mallows seemed a little better because decades of research have shown that the buildup of amyloid and tau proteins begins years, or even decades, before significant cognitive symptoms or brain atrophy are detectable~\citep{jack2010hypothetical, sperling2011toward, bateman2012clinical}.

\section{Discussion}

In this paper, we presented a framework to model the joint progression of multiple diseases. We explained how partial rankings from individual disease progressions can be used to construct a prior distribution for the aggregate rankings. 

We analyzed some theoretical properties of different JPM models. \emph{Calibration} was positive for all variants, with Mallows performing best, reflecting its distance-based energy. \emph{Separation} was near-perfect across the board, indicating that sampled rankings fall well below random permutations in energy. \emph{Sharpness} differed by variant and can be tuned (e.g., Mallows via $\theta$); importantly, \emph{sharpness} is strongly predicted by four observable properties of the input partial rankings—number of rankings, average length, \emph{conflict}, and \emph{overlap}---enabling practical variant selection before data generation or inference. 

In synthetic benchmarks, JPM delivered $21\%$ improvements in ordering accuracy over SA-EBM~\citep{hao2025stage} which treats the joint disease as a single trajectory, highlighting the advantage of incorporating multi-disease structure. In NACC, we compared all results and concluded that the results by the baseline and the Mallows variant best reflect what past literature indicated. 

More specifically, the progression of "AD only", obtained using SA-EBM, is similar to the finding by \cite{hao2025stage} which was based on the ADNI dataset: CSF biomarkers first, followed by cognitive decline and atrophy in the brain. There are differences in the two results. For example, in this study, cognitive decline seems to occur earlier than CSF biomarker changes. These differences do not alter the overall trend, though. 

Reflecting on our results, should JPM be preferred over a single-disease model (Q4) and which variant of JPM should be used for inference (Q5)? The answer is clear based on our synthetic experimental results: if data is generated by a model that has high sharpness, i.e., ground truth of the aggregate ranking is stable in repeated attempts with the same set of partial rankings, then we can expect a larger advantage of JPM over a single-disease EBM inference model. Even if such an assumption does not hold, JPM's results are not too far away from EBM, and are still reliable. Based on our synthetic experimental results, it is desirable if the inference variant matches the variant that underlies the generative process of the aggregate rankings. What this infers is that we can first estimate the \emph{sharpness} (refer to Sec.~\ref{sec:q3}) of the underlying generative model and then choose the JPM variant with a similar \emph{sharpness} for inferential tasks.

The NACC results show that the Mallows model and the SA-EBM were the top performers. This indicates that the underlying generative process of the real-world AD and VaD mixed pathology might not have a high \emph{sharpness}. Reconciling the synthetic results and real-world application is an important direction for future work. 

Although our results demonstrate that the approach has promise, there are some limitations that should be explored in future work. Firstly, we assume all patients share the same joint disease progression. Future work should explore how to incorporate subtypes such as in SuStaIn~\citep{young2018uncovering} into the JPM framework. Secondly, JPM's performance struggles when event sequences are continuous. Integrating the approach taken by the Temporal EBM~\citep{wijeratne2023temporal} could help with this issue. Lastly, some disease interactions might result in a disease progression that is completely different from the individual disease progressions. In these cases, the JPM would be inappropriate to apply. Domain knowledge of the underlying mechanisms for how the individual diseases will interact is essential for knowing when JPM is appropriate to apply and which variant to choose.

\acks{

We thank the CHTC at the University of Wisconsin-Madison for computing support. JLA was funded by the Japan Probabilistic Computing Consortium Association.

Data collection and sharing for this project was funded by the Alzheimer's Disease
Neuroimaging Initiative (ADNI) (National Institutes of Health Grant U01 AG024904) and
DOD ADNI (Department of Defense award number W81XWH-12-2-0012). ADNI is funded
by the National Institute on Aging, the National Institute of Biomedical Imaging and
Bioengineering, and through generous contributions from the following: AbbVie, Alzheimer’s
Association; Alzheimer’s Drug Discovery Foundation; Araclon Biotech; BioClinica, Inc.;
Biogen; Bristol-Myers Squibb Company; CereSpir, Inc.; Cogstate; Eisai Inc.; Elan
Pharmaceuticals, Inc.; Eli Lilly and Company; EuroImmun; F. Hoffmann-La Roche Ltd and
its affiliated company Genentech, Inc.; Fujirebio; GE Healthcare; IXICO Ltd.; Janssen
Alzheimer Immunotherapy Research \& Development, LLC.; Johnson \& Johnson
Pharmaceutical Research \& Development LLC.; Lumosity; Lundbeck; Merck \& Co., Inc.;
Meso Scale Diagnostics, LLC.; NeuroRx Research; Neurotrack Technologies; Novartis
Pharmaceuticals Corporation; Pfizer Inc.; Piramal Imaging; Servier; Takeda Pharmaceutical
Company; and Transition Therapeutics. The Canadian Institutes of Health Research is
providing funds to support ADNI clinical sites in Canada. Private sector contributions are
facilitated by the Foundation for the National Institutes of Health (www.fnih.org). The grantee
organization is the Northern California Institute for Research and Education, and the study is
coordinated by the Alzheimer’s Therapeutic Research Institute at the University of Southern
California. ADNI data are disseminated by the Laboratory for Neuro Imaging at the
University of Southern California.

The ADSP Phenotype Harmonization Consortium (ADSP-PHC) is funded by NIA (U24 AG074855, U01 AG068057 and R01 AG059716).

The NACC database is funded by NIA/NIH Grant U24 AG072122. NACC data are contributed by the NIA-funded ADRCs: P30 AG062429 (PI James Brewer, MD, PhD), P30 AG066468 (PI Oscar Lopez, MD), P30 AG062421 (PI Bradley Hyman, MD, PhD), P30 AG066509 (PI Thomas Grabowski, MD), P30 AG066514 (PI Mary Sano, PhD), P30 AG066530 (PI Helena Chui, MD), P30 AG066507 (PI Marilyn Albert, PhD), P30 AG066444 (PI David Holtzman, MD), P30 AG066518 (PI Lisa Silbert, MD, MCR), P30 AG066512 (PI Thomas Wisniewski, MD), P30 AG066462 (PI Scott Small, MD), P30 AG072979 (PI David Wolk, MD), P30 AG072972 (PI Charles DeCarli, MD), P30 AG072976 (PI Andrew Saykin, PsyD), P30 AG072975 (PI Julie A. Schneider, MD, MS), P30 AG072978 (PI Ann McKee, MD), P30 AG072977 (PI Robert Vassar, PhD), P30 AG066519 (PI Frank LaFerla, PhD), P30 AG062677 (PI Ronald Petersen, MD, PhD), P30 AG079280 (PI Jessica Langbaum, PhD), P30 AG062422 (PI Gil Rabinovici, MD), P30 AG066511 (PI Allan Levey, MD, PhD), P30 AG072946 (PI Linda Van Eldik, PhD), P30 AG062715 (PI Sanjay Asthana, MD, FRCP), P30 AG072973 (PI Russell Swerdlow, MD), P30 AG066506 (PI Glenn Smith, PhD, ABPP), P30 AG066508 (PI Stephen Strittmatter, MD, PhD), P30 AG066515 (PI Victor Henderson, MD, MS), P30 AG072947 (PI Suzanne Craft, PhD), P30 AG072931 (PI Henry Paulson, MD, PhD), P30 AG066546 (PI Sudha Seshadri, MD), P30 AG086401 (PI Erik Roberson, MD, PhD), P30 AG086404 (PI Gary Rosenberg, MD), P20 AG068082 (PI Angela Jefferson, PhD), P30 AG072958 (PI Heather Whitson, MD), P30 AG072959 (PI James Leverenz, MD).

The NACC database is funded by NIA/NIH Grant U24 AG072122. SCAN is a multi-institutional project that was funded as a U24 grant (AG067418) by the National Institute on Aging in May 2020. Data collected by SCAN and shared by NACC are contributed by the NIA-funded ADRCs as follows:

Arizona Alzheimer’s Center - P30 AG072980 (PI: Eric Reiman, MD); R01 AG069453 (PI: Eric Reiman (contact), MD); P30 AG019610 (PI: Eric Reiman, MD); and the State of Arizona which provided additional funding supporting our center; Boston University - P30 AG013846 (PI Neil Kowall MD); Cleveland ADRC - P30 AG062428 (James Leverenz, MD); Cleveland Clinic, Las Vegas – P20AG068053; Columbia - P50 AG008702 (PI Scott Small MD); Duke/UNC ADRC – P30 AG072958; Emory University - P30AG066511 (PI Levey Allan, MD, PhD); Indiana University - R01 AG19771 (PI Andrew Saykin, PsyD); P30 AG10133 (PI Andrew Saykin, PsyD); P30 AG072976 (PI Andrew Saykin, PsyD); R01 AG061788 (PI Shannon Risacher, PhD); R01 AG053993 (PI Yu-Chien Wu, MD, PhD); U01 AG057195 (PI Liana Apostolova, MD); U19 AG063911 (PI Bradley Boeve, MD); and the Indiana University Department of Radiology and Imaging Sciences; Johns Hopkins - P30 AG066507 (PI Marilyn Albert, Phd.); Mayo Clinic - P50 AG016574 (PI Ronald Petersen MD PhD); Mount Sinai - P30 AG066514 (PI Mary Sano, PhD); R01 AG054110 (PI Trey Hedden, PhD); R01 AG053509 (PI Trey Hedden, PhD); New York University - P30AG066512-01S2 (PI Thomas Wisniewski, MD); R01AG056031 (PI Ricardo Osorio, MD); R01AG056531 (PIs Ricardo Osorio, MD; Girardin Jean-Louis, PhD); Northwestern University - P30 AG013854 (PI Robert Vassar PhD); R01 AG045571 (PI Emily Rogalski, PhD); R56 AG045571, (PI Emily Rogalski, PhD); R01 AG067781, (PI Emily Rogalski, PhD); U19 AG073153, (PI Emily Rogalski, PhD); R01 DC008552, (M.-Marsel Mesulam, MD); R01 AG077444, (PIs M.-Marsel Mesulam, MD, Emily Rogalski, PhD); R01 NS075075 (PI Emily Rogalski, PhD); R01 AG056258 (PI Emily Rogalski, PhD); Oregon Health \& Science University - P30 AG066518 (PI Lisa Silbert, MD, MCR); Rush University - P30 AG010161 (PI David Bennett MD); Stanford – P30AG066515; P50 AG047366 (PI Victor Henderson MD MS); University of Alabama, Birmingham – P20; University of California, Davis - P30 AG10129 (PI Charles DeCarli, MD); P30 AG072972 (PI Charles DeCarli, MD); University of California, Irvine - P50 AG016573 (PI Frank LaFerla PhD); University of California, San Diego - P30AG062429 (PI James Brewer, MD, PhD); University of California, San Francisco - P30 AG062422 (Rabinovici, Gil D., MD); University of Kansas - P30 AG035982 (Russell Swerdlow, MD); University of Kentucky - P30 AG028283-15S1 (PIs Linda Van Eldik, PhD and Brian Gold, PhD); University of Michigan ADRC - P30AG053760 (PI Henry Paulson, MD, PhD) P30AG072931 (PI Henry Paulson, MD, PhD) Cure Alzheimer's Fund 200775 - (PI Henry Paulson, MD, PhD) U19 NS120384 (PI Charles DeCarli, MD, University of Michigan Site PI Henry Paulson, MD, PhD) R01 AG068338 (MPI Bruno Giordani, PhD, Carol Persad, PhD, Yi Murphey, PhD) S10OD026738-01 (PI Douglas Noll, PhD) R01 AG058724 (PI Benjamin Hampstead, PhD) R35 AG072262 (PI Benjamin Hampstead, PhD) W81XWH2110743 (PI Benjamin Hampstead, PhD) R01 AG073235 (PI Nancy Chiaravalloti, University of Michigan Site PI Benjamin Hampstead, PhD) 1I01RX001534 (PI Benjamin Hampstead, PhD) IRX001381 (PI Benjamin Hampstead, PhD); University of New Mexico - P20 AG068077 (Gary Rosenberg, MD); University of Pennsylvania - State of PA project 2019NF4100087335 (PI David Wolk, MD); Rooney Family Research Fund (PI David Wolk, MD); R01 AG055005 (PI David Wolk, MD); University of Pittsburgh - P50 AG005133 (PI Oscar Lopez MD); University of Southern California - P50 AG005142 (PI Helena Chui MD); University of Washington - P50 AG005136 (PI Thomas Grabowski MD); University of Wisconsin - P50 AG033514 (PI Sanjay Asthana MD FRCP); Vanderbilt University – P20 AG068082; Wake Forest - P30AG072947 (PI Suzanne Craft, PhD); Washington University, St. Louis - P01 AG03991 (PI John Morris MD); P01 AG026276 (PI John Morris MD); P20 MH071616 (PI Dan Marcus); P30 AG066444 (PI John Morris MD); P30 NS098577 (PI Dan Marcus); R01 AG021910 (PI Randy Buckner); R01 AG043434 (PI Catherine Roe); R01 EB009352 (PI Dan Marcus); UL1 TR000448 (PI Brad Evanoff); U24 RR021382 (PI Bruce Rosen); Avid Radiopharmaceuticals / Eli Lilly; Yale - P50 AG047270 (PI Stephen Strittmatter MD PhD); R01AG052560 (MPI: Christopher van Dyck, MD; Richard Carson, PhD); R01AG062276 (PI: Christopher van Dyck, MD); 1Florida - P30AG066506-03 (PI Glenn Smith, PhD); P50 AG047266 (PI Todd Golde MD PhD)

}

\bibliography{paper}

\onecolumn
\appendix

\clearpage 

\section{ADNI Information}
\label{sec:adni}


A complete listing of ADNI
investigators can be found at:
\url{http://adni.loni.usc.edu/wp-content/uploads/how_to_apply/ADNI_Acknowledgement_List.pdf}

\section{Biomarkers Involved in each disease}
\label{apd:biomarkers_disease}

\begin{itemize}
    \item AD: ABETA, PTAU, TAU, Entorhinal, Hippocampus, Fusiform, MidTemp, WholeBrain, FDG, ADAS13, CDRSB, LDELTOTAL, RAVLT, MMSE, MOCA, TRABSCOR, FAQ
    \item Vascular dementia: FDG, WholeBrain, TRABSCOR, MMSE, MOCA, FAQ, ADAS13, CDRSB
    \item LBD: FDG, Hippocampus, Entorhinal, ABETA, TAU, PTAU, MMSE, MOCA, TRABSCOR, FAQ
\end{itemize}

\section{Sampling Algorithms}
\label{apd:sampling}


\begin{algorithm2e}[htbp]
\label{alg:jpm_algo}
\caption{\textsc{JPM} Inference Algorithm}
\KwIn{Individual diseases $\mathcal{D} = \{D^{(1)}, \dots, D^{(K)}\}$, mixed-pathology data $D_{m-p}$, energy function $E(\sigma)$, data likelihood function $P(D_{m-p} \mid \sigma)$}
\KwOut{Best aggregate ranking $\sigma^\star$}

\tcp{Step 0: Estimate partial rankings from individual diseases}
Estimate partial rankings $\mathcal{S}$ from $\mathcal{D}$\;

\tcp{Init}
$\sigma \gets$ random permutation of $\mathcal{X}$\;
$\ell \gets -\infty$\;
$\sigma^\star \gets \sigma$, $\ell^\star \gets \ell$\;

\For{$t = 1$ \KwTo $T$}{
  \tcp{Step 1: Proposal}
  Propose $\sigma' \gets$ swap two random items in $\sigma$\;
  
  \tcp{Step 2: Compute data likelihood}
  $\ell' \gets \log P(D_c \mid \sigma') - E(\sigma')$\;
  
  \tcp{Step 3: Acceptance probability}
  $p \gets \min(1, \exp(\ell' - \ell))$\;
  Draw $u \sim \text{Uniform}(0,1)$\;
  
  \tcp{Step 4: Accept/Reject}
  \If{$u < p$}{
    $\sigma \gets \sigma'$,
    $\ell \gets \ell'$\;
  }
  
  \tcp{Step 5: Track best result}
  \If{$\ell > \ell^\star$}{
    $\sigma^\star \gets \sigma$,
    $\ell^\star \gets \ell$\;
  }
}

\Return $\sigma^\star$\;

\end{algorithm2e}

Explanations:

\begin{itemize}
    \item The energy function is one of the required inputs. This means we need to decide on the JPM variant. 
    \item We use EBM as the data likelihood function, i.e., $P(D_{m-p}\mid \sigma)$ and $\log P(D_{m-p}\mid \sigma)$. More specifically, we used SA-EBM~\citep{hao2025stage}. Note that EBM is just one choice for the likelihood function. Future work can explore other possible likelihoods.
    \item In the estimation of partial rankings from individual diseases' data, we also used SA-EBM~\citep{hao2025stage}, which has state-of-the-art performance on estimating ordinal disease progressions from cross-sectional datasets. 
\end{itemize}



\begin{algorithm2e}[htbp]
\label{alg:pl_sample}
\caption{Plackett--Luce Sampling}
\KwIn{Parameters $\vec{\alpha} = (\alpha_1, \dots, \alpha_m)$}
\KwOut{Aggregate ranking $\sigma = [i_1, i_2, \dots, i_m]$}

\For{$t = 1$ \KwTo $m$}{
  \tcp{Sample next item proportional to its weight}
  \tcp{$\mathcal{X}$ is the list of biomarkers in the aggregate ranking, whose length is $m$}
  Sample $i_t$ with probability 
  $\displaystyle \frac{e^{\alpha_i}}{\sum_{q \in \mathcal{X}} e^{\alpha_q}}$\;
  
  \tcp{Remove item from available set}
  $U \gets \mathcal{X} \setminus \{i_t\}$\;
}

\Return $\sigma = [i_1, i_2, \dots, i_m]$\;

\end{algorithm2e}


\begin{algorithm2e}[htbp]
\label{alg:mh_sample}
\caption{\textsc{JMP} Generative Algorithm (Metropolis--Hastings MCMC)}
\KwIn{Set $\mathcal{X}$, Energy function $E(\sigma)$}
\KwOut{Best aggregate ranking $\sigma^\star$}

\tcp{Init}
$\sigma \gets$ random permutation of $\mathcal{X}$\;
$\ell \gets E(\sigma)$\;
$\sigma^\star \gets \sigma$, $\ell^\star \gets \ell$\;

\For{$t = 1$ \KwTo $T$}{
  \tcp{Step 1: Proposal}
  Propose $\sigma' \gets$ swap two random items in $\sigma$\;
  
  \tcp{Step 2: Compute energy}
  $\ell' \gets E(\sigma')$\;
  
  \tcp{Step 3: Metropolis-Hastings acceptance probability}
  $p \gets \min(1, \exp(\ell - \ell'))$\;
  Draw $u \sim \text{Uniform}(0,1)$\;
  
  \tcp{Step 4: Accept/Reject}
  \If{$u < p$}{
    $\sigma \gets \sigma'$, $\ell \gets \ell'$\;
  }
  
  \tcp{Step 5: Track minimum energy state}
  \If{$\ell < \ell^\star$}{
    $\sigma^\star \gets \sigma$, $\ell^\star \gets \ell$\;
  }
}

\Return $\sigma^\star$\;

\end{algorithm2e}

\clearpage
\section{Normalized Kendall's $\tau$ distance}
\label{apd:normalized_tau_distance}

\begin{algorithm2e}[htbp]
\label{alg:normalized_tau_distance}
\caption{Normalized Kendall's Tau Distance}
\KwIn{Two rankings $r_1, r_2$ of length $n$. The two rankings are two arrays of indices of the same set of items. }
\KwOut{Normalized Kendall's tau distance $d \in [0,1]$}

$concordant \gets 0$, $discordant \gets 0$\;

\For{$p = 1$ \KwTo $n-1$}{
  \For{$q = p+1$ \KwTo $n$}{
    \tcp{Compare relative order of items $p$ and $q$ in both rankings}
    \If{$(r_1[p] - r_1[q]) \cdot (r_2[p] - r_2[q]) > 0$}{
      $concordant \gets concordant + 1$\;
    }
    \If{$(r_1[p] - r_1[q]) \cdot (r_2[p] - r_2[q]) < 0$}{
      $discordant \gets discordant + 1$\;
    }
  }
}

$total \gets concordant + discordant$\;

\If{$total = 0$}{
  \Return $0$\;
}
\Else{
  \Return $\displaystyle \frac{discordant}{total}$\;
}
\end{algorithm2e}

\clearpage
\section{Model Analysis}
\label{apd:model_analysis}

\begin{table*}[htbp]
\caption{Regression Models Predicting Model Sharpness}
\label{tab:combined_sharpness}
\centering
\begin{tabular}{lccccc}
\toprule
 & \multicolumn{1}{c}{PL} & \multicolumn{1}{c}{BT} & \multicolumn{1}{c}{Pairwise} & \multicolumn{1}{c}{Mallows $\theta=1.0$} & \multicolumn{1}{c}{Mallows $\theta=10.0$} \\
 & (1) & (2) & (3) & (4) & (5) \\
\midrule
const & 1.142*** & 1.013*** & 0.942*** & 1.082*** & 1.240*** \\
 & (0.015) & (0.002) & (0.010) & (0.005) & (0.005) \\
$n_{pr}$ & -0.003 & -0.000 & 0.001 & -0.113*** & -0.121*** \\
 & (0.002) & (0.000) & (0.001) & (0.001) & (0.001) \\
mean\_len & 0.004*** & 0.001*** & 0.002** & -0.047*** & -0.048*** \\
 & (0.001) & (0.000) & (0.001) & (0.000) & (0.001) \\
conflict & -0.257*** & -0.030*** & -0.116*** & -0.001 & 0.002 \\
 & (0.008) & (0.001) & (0.006) & (0.002) & (0.003) \\
overlap\_rate & -0.567*** & -0.073*** & -0.095*** & 0.283*** & 0.292*** \\
 & (0.008) & (0.001) & (0.006) & (0.005) & (0.005) \\
\midrule
$R^2$ & 0.651 & 0.630 & 0.193 & 0.858 & 0.852 \\
Adj. $R^2$ & 0.650 & 0.629 & 0.192 & 0.858 & 0.851 \\
$N$ & 3240 & 3240 & 3240 & 3240 & 3240 \\
\bottomrule
\multicolumn{6}{l}{\footnotesize Standard errors in parentheses. * $p$$<$0.1, ** $p$$<$0.05, *** $p$$<$0.01} \\
\end{tabular}
\end{table*}

In Table~\ref{tab:combined_sharpness}, $n_{pr}$ refers to the number of partial rankings, and mean\_len indicates the average number of partial rankings.

\begin{figure*}[htbp]
    \centering \includegraphics[width=1.0\linewidth]{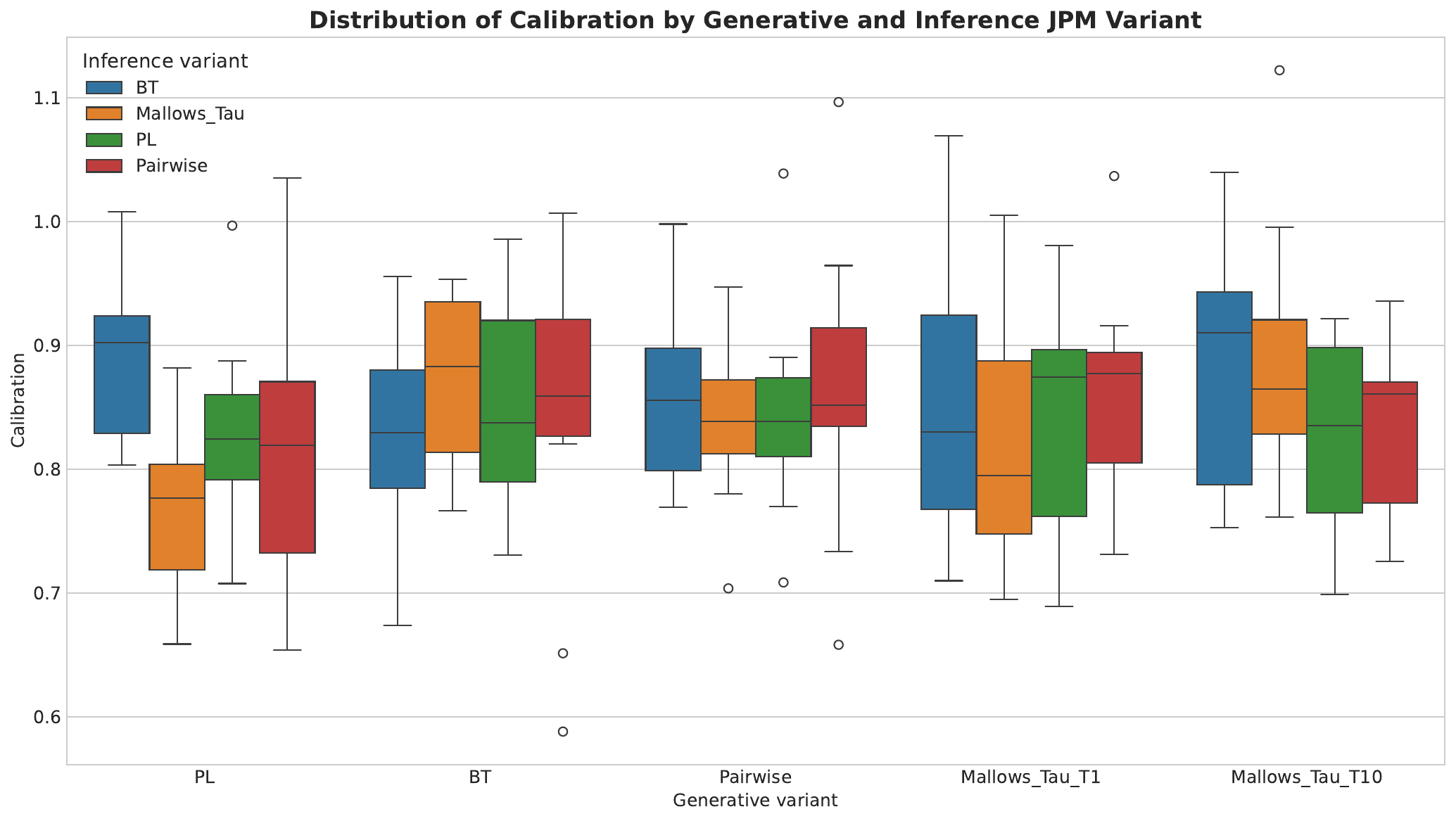} 
    \vspace{-2em}
    \caption{Distribution of calibration by generative and inference JPM variant}
    \label{fig:clb} 
\end{figure*}

\begin{figure*}[htbp]
    \centering \includegraphics[width=1.0\linewidth]{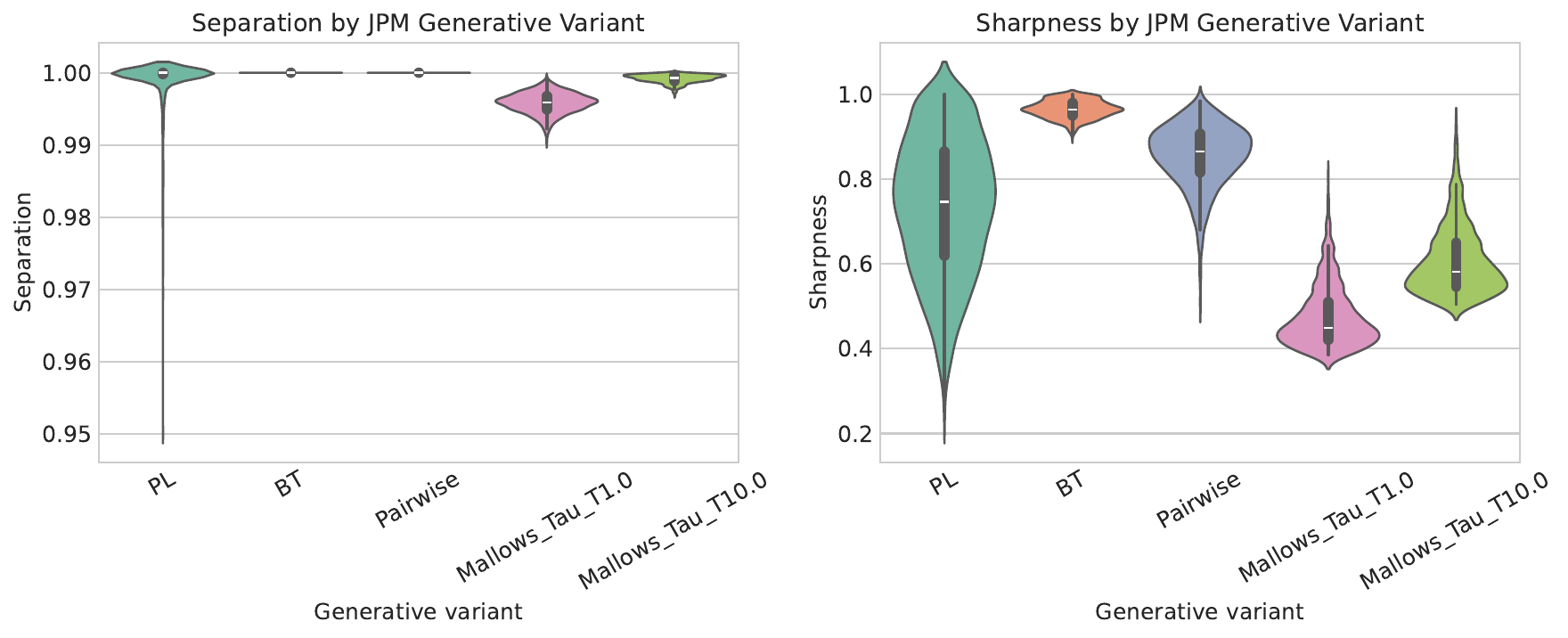} 
    \vspace{-2em}
    \caption{Separation and sharpness of different JPM variants}
    \label{fig:sep} 
\end{figure*}

\begin{figure*}[htbp]
    \centering \includegraphics[width=1.0\linewidth]{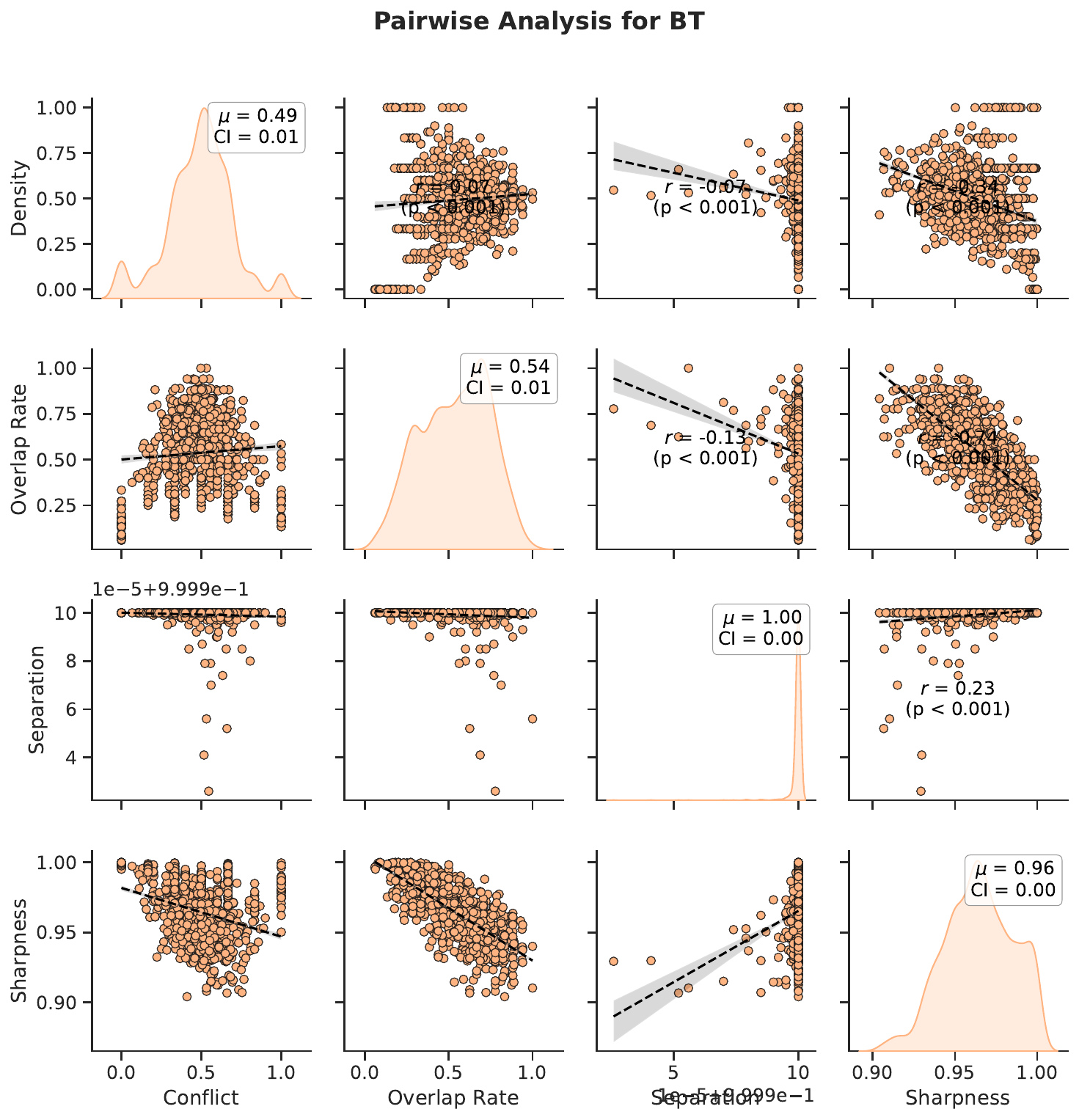} 
    \vspace{-2em}
    \caption{Correlation between sharpness and input partial ranking characteristics when the aggregate rankings are generated with the BT variant}
    \label{fig:bt} 
\end{figure*}

\begin{figure*}[htbp]
    \centering \includegraphics[width=1.0\linewidth]{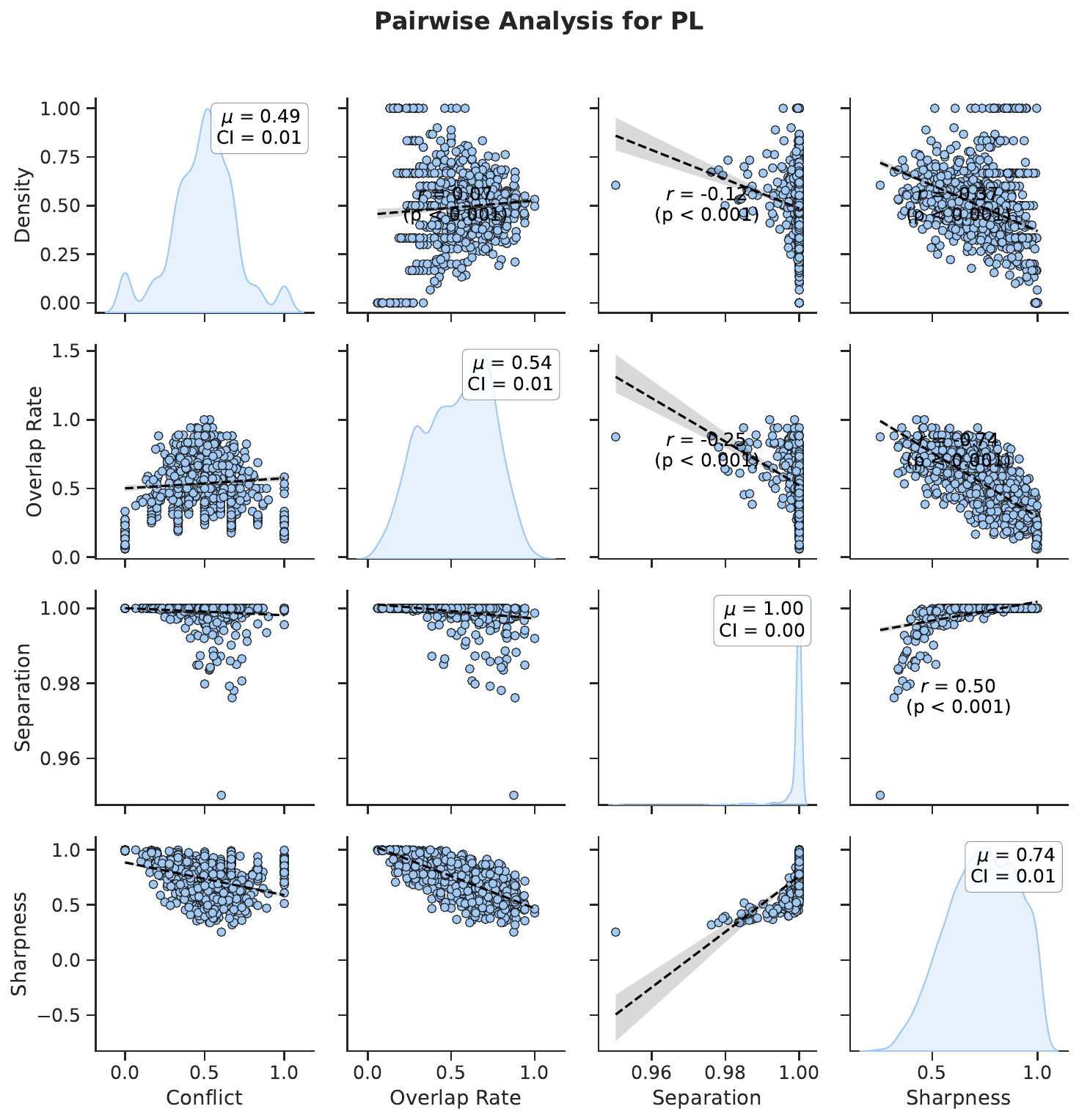} 
    \vspace{-2em}
    \caption{Correlation between sharpness and input partial ranking characteristics when the aggregate rankings are generated with the PL variant}
    \label{fig:pl} 
\end{figure*}

\begin{figure*}[htbp]
    \centering \includegraphics[width=1.0\linewidth]{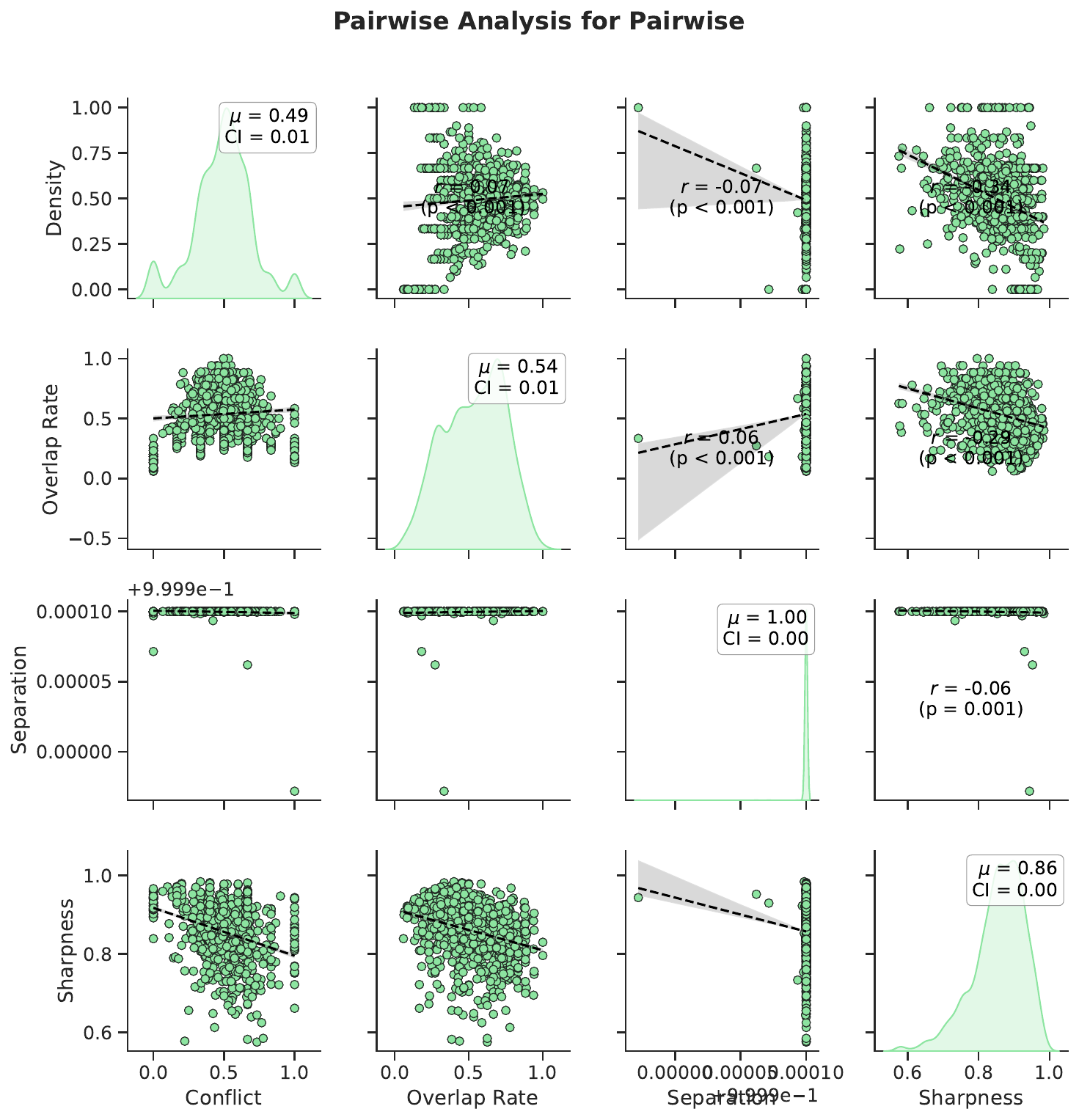} 
    \vspace{-2em}
    \caption{Correlation between sharpness and input partial ranking characteristics when the aggregate rankings are generated with the PP variant}
    \label{fig:pp} 
\end{figure*}

\begin{figure*}[htbp]
    \centering \includegraphics[width=1.0\linewidth]{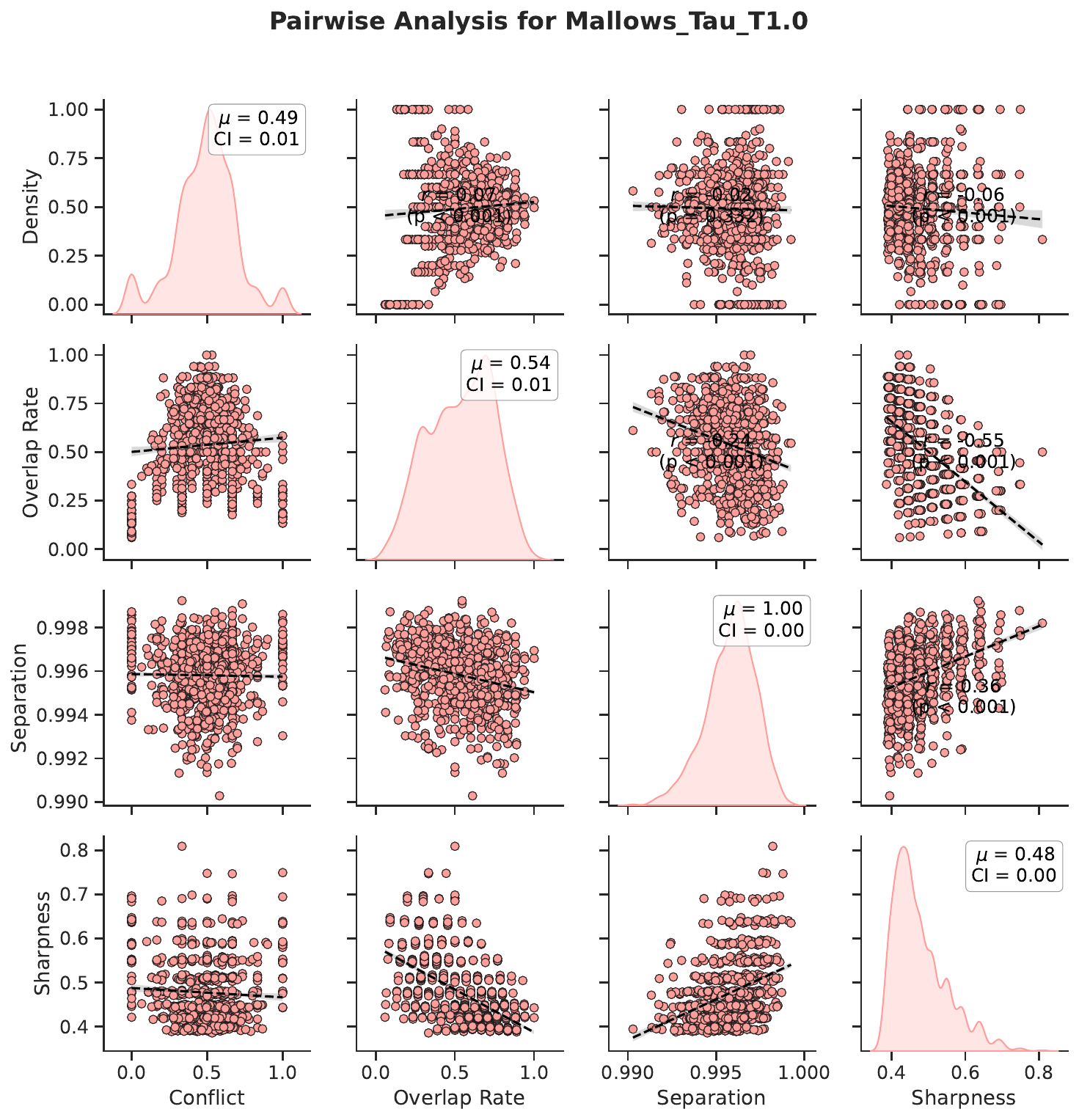} 
    \vspace{-2em}
    \caption{Correlation between sharpness and input partial ranking characteristics when the aggregate rankings are generated with the Mallows ($\theta=1$) variant}
    \label{fig:m1} 
\end{figure*}

\begin{figure*}[htbp]
    \centering \includegraphics[width=1.0\linewidth]{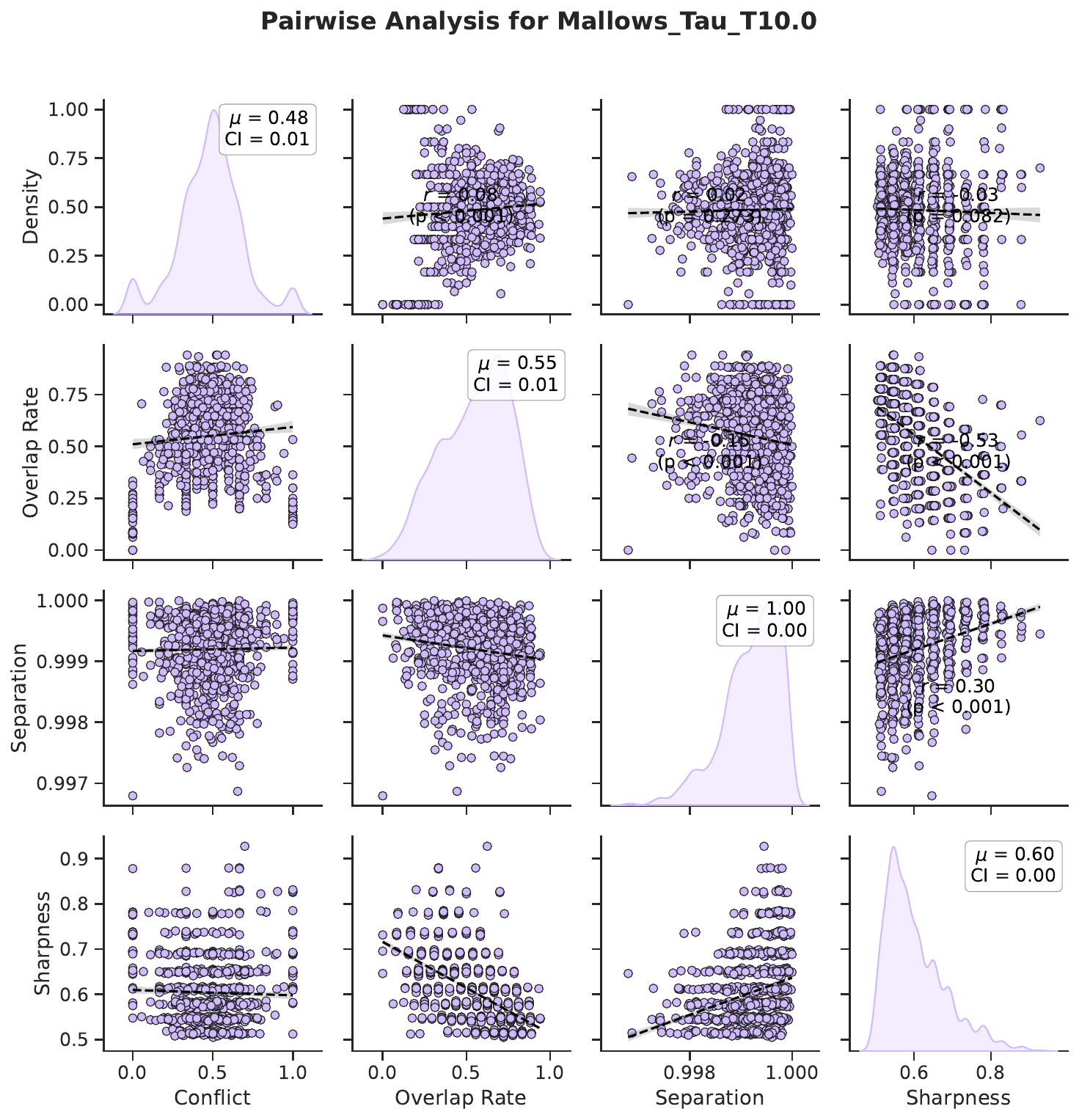} 
    \vspace{-2em}
    \caption{Correlation between sharpness and input partial ranking characteristics when the aggregate rankings are generated with the Mallows ($\theta=10$) variant}
    \label{fig:m10} 
\end{figure*}

\clearpage 
\section{Biomarkers for synthetic experiments}
\label{apd:biomarkers}

This is how we processed the ADNI data and obtained the parameters reported in Table~\ref{tab:biomarker_params}. We obtained the \texttt{adnimerge} table (Version: September 7, 2023) from the Alzheimer's Disease Cooperative Study data system. We only included the baseline visits from participants with a diagnosis of CN, Early and Late MCI, or AD. For brain region MRI values, we applied intracranial volume (ICV) normalization because people have different brain sizes. We excluded participants with any missing values from any of the 18 biomarkers reported in Table~\ref{tab:biomarker_glossary_extended}. After de-duplicating the final dataset,  we had 420 participants with 350 (83.3\%) from the protocol of ADNI2 and 70 (18.3\%) from ADNIGO. The distribution of diagnosis of these 420 participants is:

\begin{itemize}
    \item AD: 67 (16.0\%)
    \item EMCI: 175 (41.7\%)
    \item LMCI: 101 (24.0\%)
    \item CN: 77 (18.3\%)
\end{itemize}

We then applied SA-EBM~\citep{hao2025stage} (the Conjugate Priors variant) to obtain the biomarker distribution parameters reported in Table~\ref{tab:biomarker_params}.

\begin{table*}[ht]
\centering
\caption{Glossary of Biomarkers with Source, Units, and Interpretation}
\resizebox{\linewidth}{!}{%
\label{tab:biomarker_glossary_extended}
\begin{tabular}{lllll}
\hline
\textbf{Abbrev.} & \textbf{Full Name} & \textbf{Source} & \textbf{Unit / Scale} & \textbf{Higher Values Indicate} \\
\hline
ABETA        & Amyloid Beta (A$\beta_{1-42}$) & CSF & pg/mL & Less pathology (less amyloid) \\
ADAS13       & ADAS-Cog (13-item) & Cognitive test & 0--85 & More pathology (worse cognition) \\
CDRSB        & CDR Sum of Boxes & Clinical rating & 0--18 & More pathology (greater severity) \\
EntorhinalNorm & Entorhinal Volume (norm.) & MRI & Ratio & Less pathology (greater volume) \\
FAQ          & Functional Activities Q. & Informant test & 0--30 & More pathology (worse function) \\
FDG          & FDG-PET (PCC/precuneus) & PET & SUVR & Less pathology (better metabolism) \\
FusiformNorm & Fusiform Volume (norm.) & MRI & Ratio & Less pathology (greater volume) \\
HippocampusNorm & Hippocampal Volume (norm.) & MRI & Ratio & Less pathology (greater volume) \\
LDELTOTAL    & Logical Memory II (Delayed) & Cognitive test & 0--25 & Less pathology (better recall) \\
MMSE         & Mini-Mental State Exam & Cognitive test & 0--30 & Less pathology (better cognition) \\
MOCA         & Montreal Cognitive Assessment & Cognitive test & 0--30 & Less pathology (better cognition) \\
MidTempNorm  & Middle Temporal Volume (norm.) & MRI & Ratio & Less pathology (greater volume) \\
PTAU         & Phosphorylated Tau$_{181}$ & CSF & pg/mL & More pathology (tangle burden) \\
RAVLT-immediate & Rey Auditory Verbal Learning (Immediate) & Cognitive test & 0--75 & Less pathology (better memory) \\
TAU          & Total Tau & CSF & pg/mL & More pathology (axonal injury) \\
TRABSCOR     & Trail Making Test B & Cognitive test & Seconds & More pathology (slower exec. func.) \\
VentricleNorm & Ventricular Volume (norm.) & MRI & Ratio & More pathology (ventricular enlarge.) \\
WholeBrainNorm & Whole Brain Volume (norm.) & MRI & Ratio & Less pathology (greater volume) \\
\hline
\end{tabular}
}
\end{table*}

\begin{table}[ht]
\centering
\caption{Biomarker Parameterization and Non-Normal Sampling Distributions}
\label{tab:biomarker_params}
\scriptsize
\begin{tabular}{lccccp{6cm}}
\toprule
\textbf{Biomarker} & $\theta_{\text{mean}}$ & $\theta_{\text{std}}$ & $\phi_{\text{mean}}$ & $\phi_{\text{std}}$ & \textbf{Non-normal Distribution (Per Implementation)} \\
\midrule

\textbf{Cognitive / Functional Measures} \\
MMSE & 24.74 & 2.26 & 28.75 & 1.30 &
Triangular$(\mu-2\sigma, \mu-1.5\sigma, \mu)$;
$\mathcal{N}(\mu+\sigma, (0.3\sigma)^2)$;
Exp$(0.7\sigma)+(\mu-0.5\sigma)$ (equal mixture). \\

ADAS13 & 26.09 & 7.77 & 11.22 & 4.83 &
Same mixture structure as MMSE (triangular + Gaussian + exponential). \\

RAVLT\_immediate & 25.00 & 6.39 & 41.70 & 10.42 &
Same mixture structure as MMSE. \\

CDRSB & 2.12 & 1.66 & 0.04 & 0.16 &
Same mixture structure as MMSE. \\

FAQ & 8.12 & 6.54 & 0.52 & 0.90 &
Same mixture structure as MMSE. \\

LDELTOTAL & 2.95 & 2.82 & 9.77 & 3.53 &
Same mixture structure as MMSE. \\

MOCA & 18.85 & 3.52 & 24.46 & 2.62 &
Same mixture structure as MMSE. \\

TRABSCOR & 276.62 & 32.45 & 97.77 & 43.00 &
Same mixture structure as MMSE. \\
\midrule

\textbf{CSF Biomarkers \& FDG} \\
ABETA & 712.35 & 240.58 & 1077.89 & 353.53 &
Pareto(1.5)$\cdot\sigma+(\mu-2\sigma)$;
$\mathcal{U}(\mu-1.5\sigma,\mu+1.5\sigma)$;
Logistic$(\mu,\sigma)$ (equal mixture). \\

TAU & 350.36 & 148.44 & 211.02 & 78.13 &
Same mixture structure as ABETA. \\

PTAU & 34.99 & 16.15 & 19.46 & 8.24 &
Same mixture structure as ABETA. \\

FDG & 1.09 & 0.12 & 1.29 & 0.12 &
Same mixture structure as ABETA. \\
\midrule

\textbf{Subcortical Volumes} \\
VentricleNorm & 0.03256 & 0.01228 & 0.02064 & 0.00929 &
Beta(0.5,0.5)$\cdot4\sigma+(\mu-2\sigma)$;
Exp$(0.4\sigma)$ with $\pm$ sign;
$\mathcal{N}(\mu,(0.5\sigma)^2)$ + \{0, $2\sigma$\} spike. \\

HippocampusNorm & 0.00394 & 0.00053 & 0.00502 & 0.00059 &
Same mixture structure as VentricleNorm. \\
\midrule

\textbf{Cortical Atrophy Measures} \\
WholeBrainNorm & 0.66854 & 0.03340 & 0.71070 & 0.03502 &
Gamma$(2, 0.5\sigma)+(\mu-\sigma)$;
Weibull$(1.0)\cdot\sigma+(\mu-\sigma)$;
$\mathcal{N}(\mu,(0.5\sigma)^2)\pm\sigma$. \\

EntorhinalNorm & 0.001997 & 0.000401 & 0.002576 & 0.000377 &
Same mixture structure as WholeBrainNorm. \\

FusiformNorm & 0.01095 & 0.00127 & 0.01255 & 0.00131 &
Standard Cauchy$(\mu,\sigma)$ + $\mathcal{N}(0,(0.2\sigma)^2)$, clipped to $[\mu-4\sigma,\mu+4\sigma]$. \\

MidTempNorm & 0.01205 & 0.00145 & 0.01397 & 0.00129 &
10\% $\mathcal{N}(\mu,0.2\sigma)$ spike + 90\% Logistic$(\mu+\sigma,2\sigma)$. \\
\bottomrule
\end{tabular}

\vspace{1mm}
\raggedright
\small
\textbf{Implementation Notes:} After sampling, all values are perturbed by additional noise $\mathcal{N}(0,(0.2\sigma)^2)$ and clipped to $[\mu-5\sigma,\mu+5\sigma]$ for stability.
\end{table}

\clearpage
\section{Experimental setup}
\label{apd:experimental_setup}
\begin{table*}[ht]
\centering
\caption{Summary of nine simulation experiments. $S$ denotes the event sequence, $k_j$ the participant stage distribution, and $X$ the biomarker distribution.}
\resizebox{\linewidth}{!}{%
\label{tab:exp_configs}
\begin{tabular}{clll}
\hline
\textbf{Exp.} & \textbf{Event Sequence ($S$)} & \textbf{Stage Distribution ($k_j$)} & \textbf{Biomarker Distribution ($X$)} \\
\hline
1 & Ordinal & Dirichlet-Multinomial (bell-shaped) & Normal \\
2 & Ordinal & Dirichlet-Multinomial (bell-shaped) & Non-Normal \\
3 & Ordinal & Uniform (discrete)                  & Normal \\
4 & Ordinal & Uniform (discrete)                  & Non-Normal \\
5 & Ordinal & Continuous Uniform                  & Normal \\
6 & Ordinal & Continuous Uniform                  & Non-Normal \\
7 & Ordinal & Continuous Skewed (Beta, $\alpha=5,\;\beta=2$) & Non-Normal \\
8 & Ordinal & Continuous Uniform                  & Sigmoid for diseased; Normal for healthy \\
9 & Ordinal & Continuous Skewed (Beta, $\alpha=5,\;\beta=2$) & Sigmoid for diseased; Normal for healthy \\
\hline
\end{tabular}
}
\end{table*}

\clearpage 
\section{Sigmoid Model}
\label{apd:sigmoid}

The \textbf{Sigmoid model} assumes that biomarker values for control individuals are normally distributed:
\[
x \sim \mathcal{N}(\mu_{\phi}, \sigma^2_{\phi}),
\]
where $\mu_{\phi}$ and $\sigma^2_{\phi}$ denote the mean and variance in the healthy state, respectively. 

For participants affected by disease, the biomarker distribution is shifted away from this baseline according to a smooth, sigmoidal transition:
\[
x \sim \mathcal{N}(\mu_{n,\phi}, \sigma^2_{n,\phi}) \;+\; \frac{(-1)^{I_n} R_n}{1 + \exp\!\left(-\rho_n (k_j - \xi_n)\right)},
\]
where $I_n \sim \text{Bernoulli}(0.5)$ randomly determines the direction of the shift, $R_n = \mu_{n,\theta} - \mu_{n,\phi}$ specifies the magnitude of separation between disease and control means, and the slope parameter is given by
\[
\rho_n = \max\!\left(1,\; \frac{|R_n|}{\sqrt{\sigma^2_{n,\theta} + \sigma^2_{n,\phi}}}\right).
\]

\clearpage
\section{Synthetic experiment results}
\label{apd:synthetic_results}

\begin{figure*}[htbp]
    \centering \includegraphics[width=1.0\linewidth]{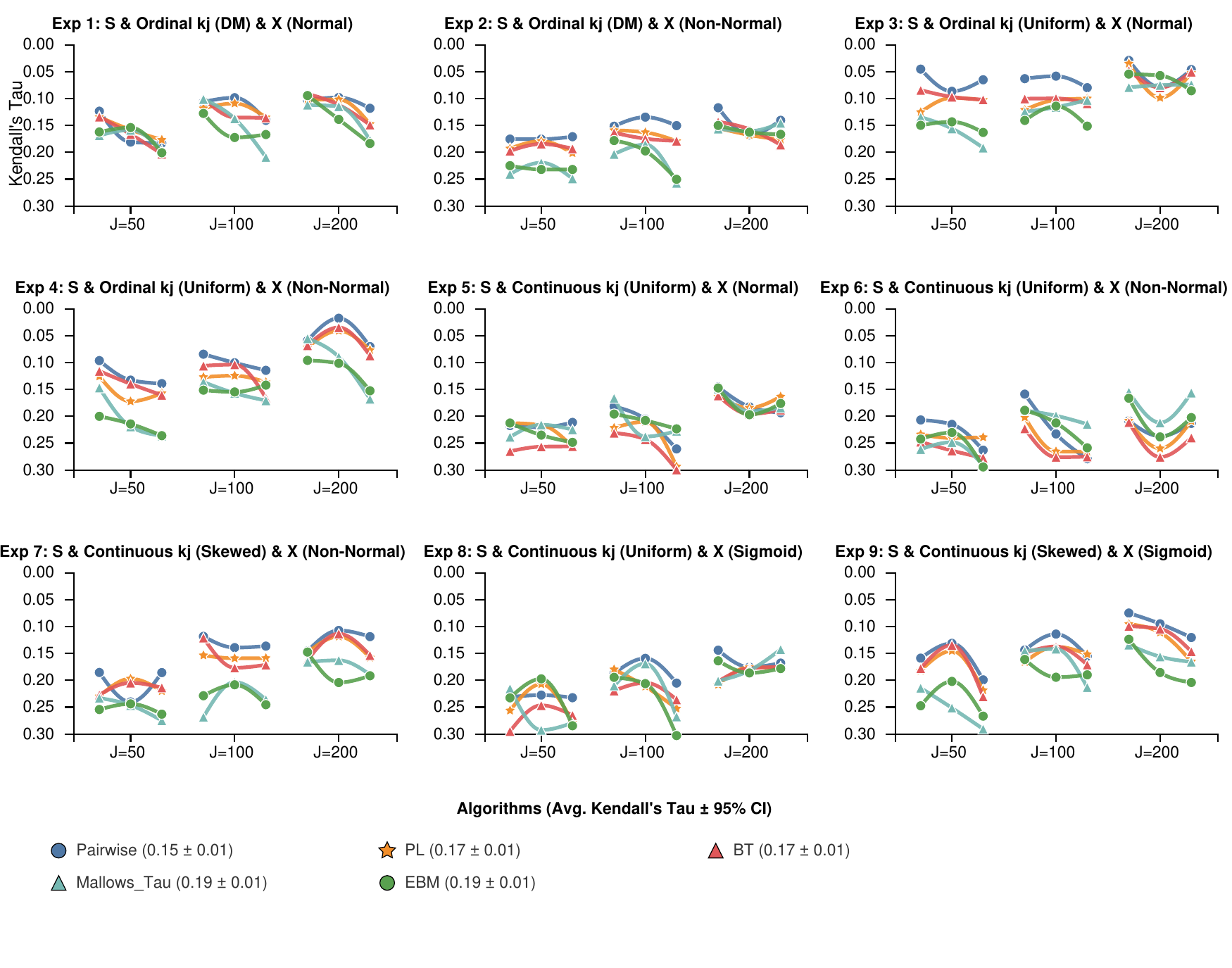} 
    \vspace{-2em}
    \caption{\textbf{Average normalized Kendall's Tau ($\pm 95\%$ CI) for data generated by PP}. Each panel displays a different experiment with varying configurations represented by the panel title. The X-axis of each subplot shows participant sizes ($J = 50, 100, 200$). For each participant size, there are three different healthy ratios ($0.25, 0.5, 0.75$, from left to right). The Y-axis is the normalized Kendall's tau distance. A lower tau indicates the estimation is closer to the ground truth. Data points represent mean performance across 10 variants of the same experimental configurations, participant size, and healthy ratio. JPM (except for Mallows with $\theta=1.0$) outperforms the single-disease EBM model, achieving a 21\% increase in the performance of the ordering task. This advantage is consistent across experiments, participant sizes, and healthy ratios, but more pronounced when $J$ is small.}
    \label{fig:pp_tau} 
\end{figure*}

\begin{figure*}[htbp]
    \centering \includegraphics[width=1.0\linewidth]{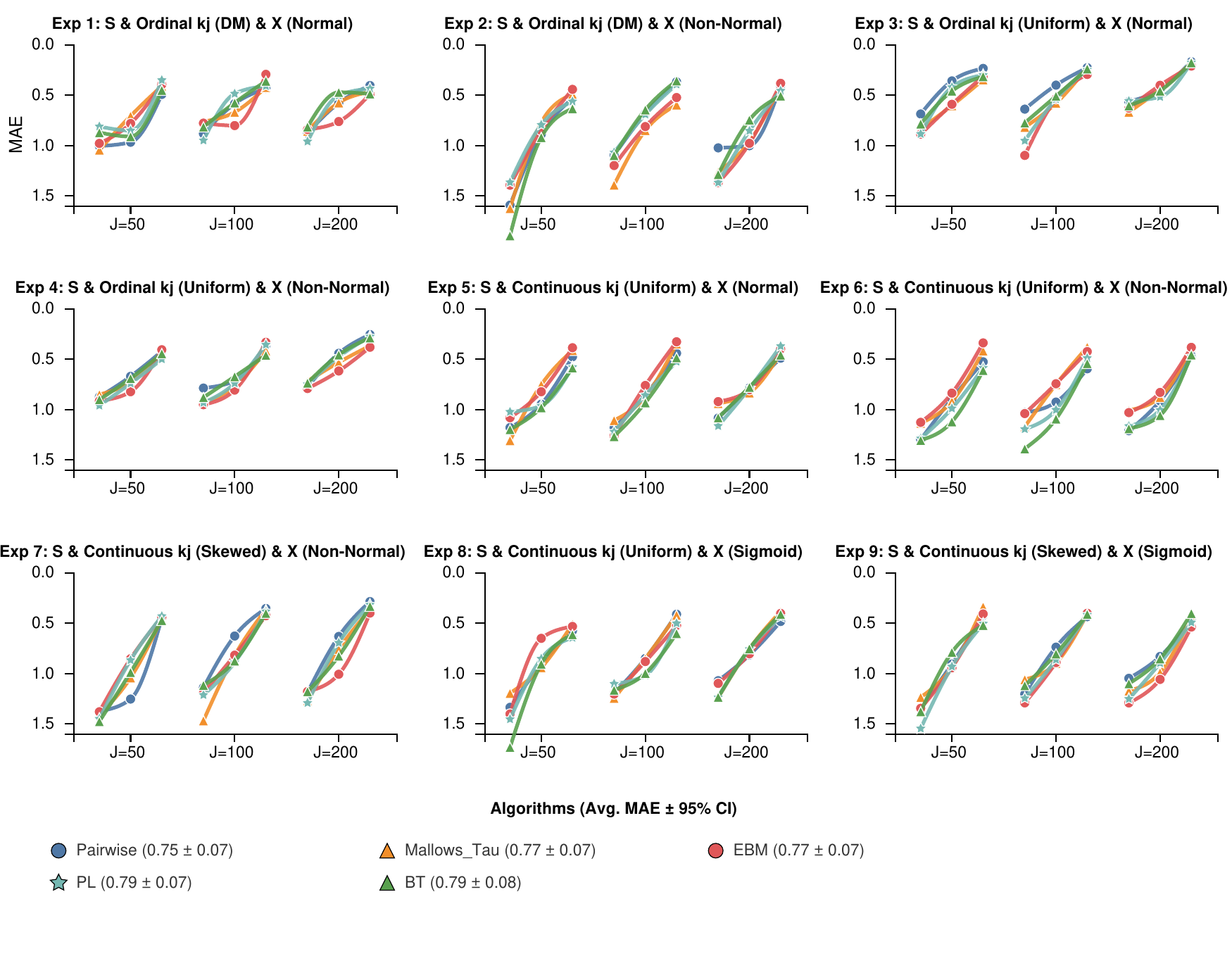} 
    \vspace{-2em}
    \caption{MAE (staging) for PP generated data
    }
    \label{fig:pp_mae} 
\end{figure*}

\begin{figure*}[htbp]
    \centering \includegraphics[width=1.0\linewidth]{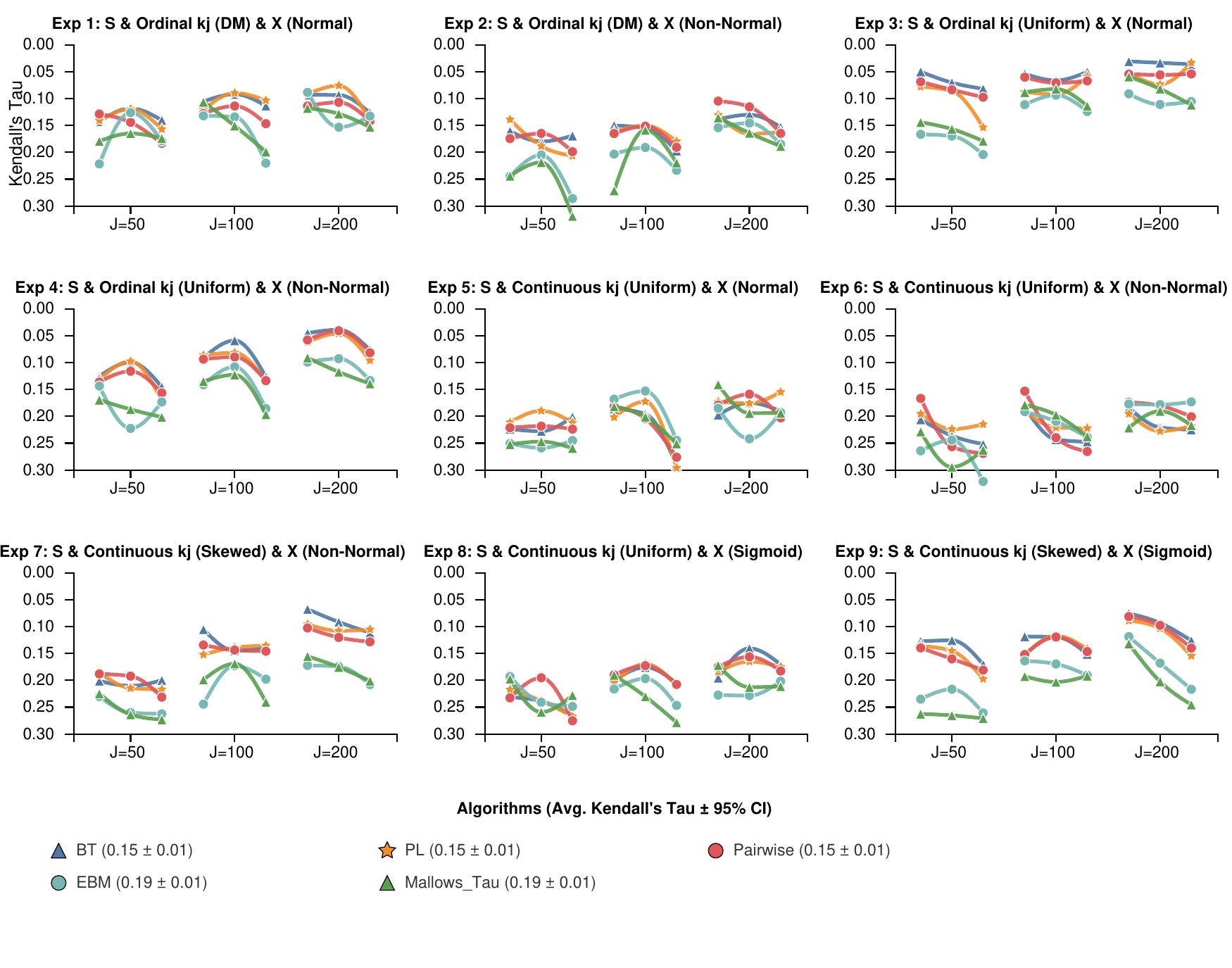} 
    \vspace{-2em}
    \caption{Kendall's tau distance (ordering) for BT generated data
    }
    \label{fig:bt_tau} 
\end{figure*}

\begin{figure*}[htbp]
    \centering \includegraphics[width=1.0\linewidth]{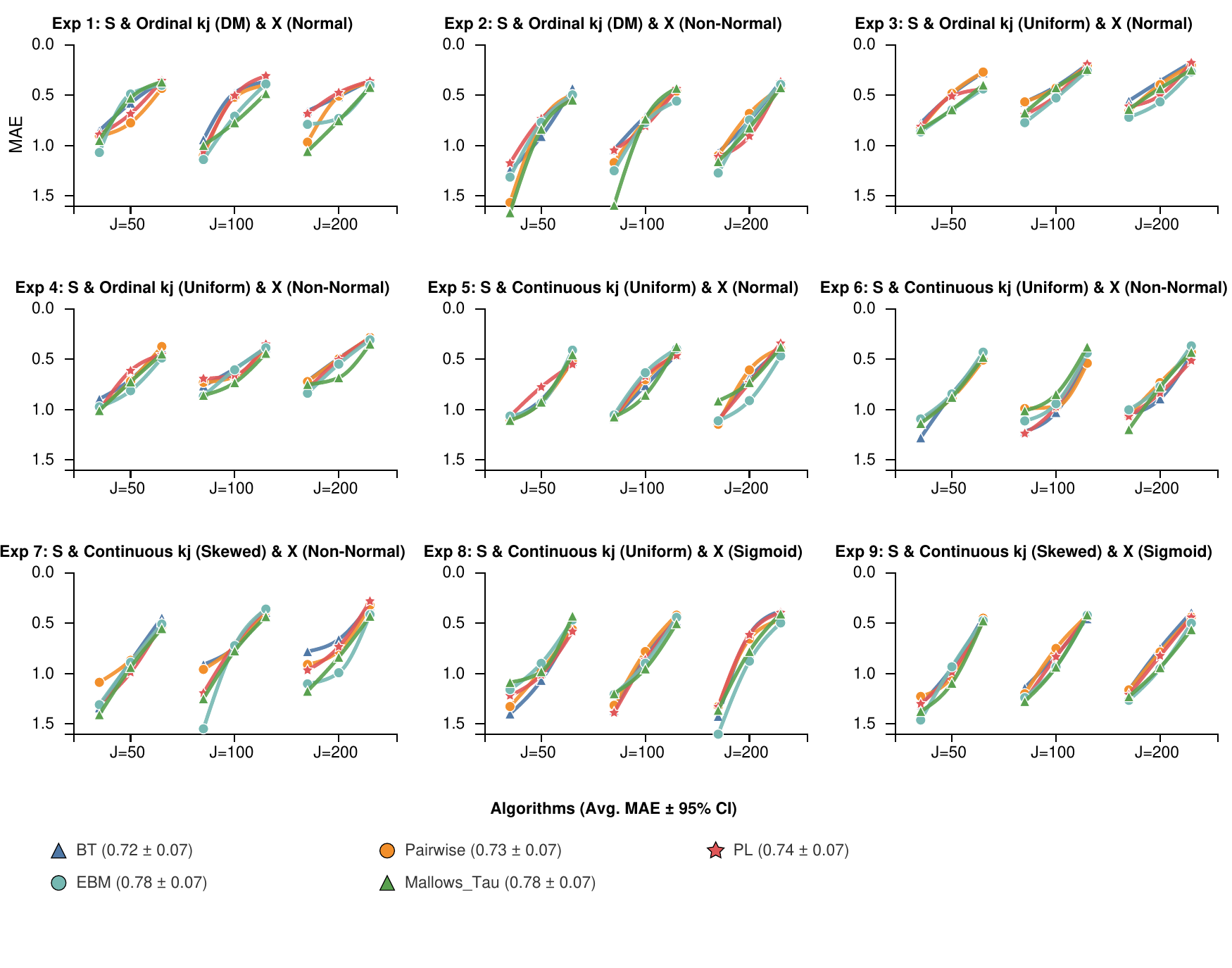} 
    \vspace{-2em}
    \caption{MAE (staging) for BT generated data
    }
    \label{fig:bt_mae} 
\end{figure*}

\begin{figure*}[htbp]
    \centering \includegraphics[width=1.0\linewidth]{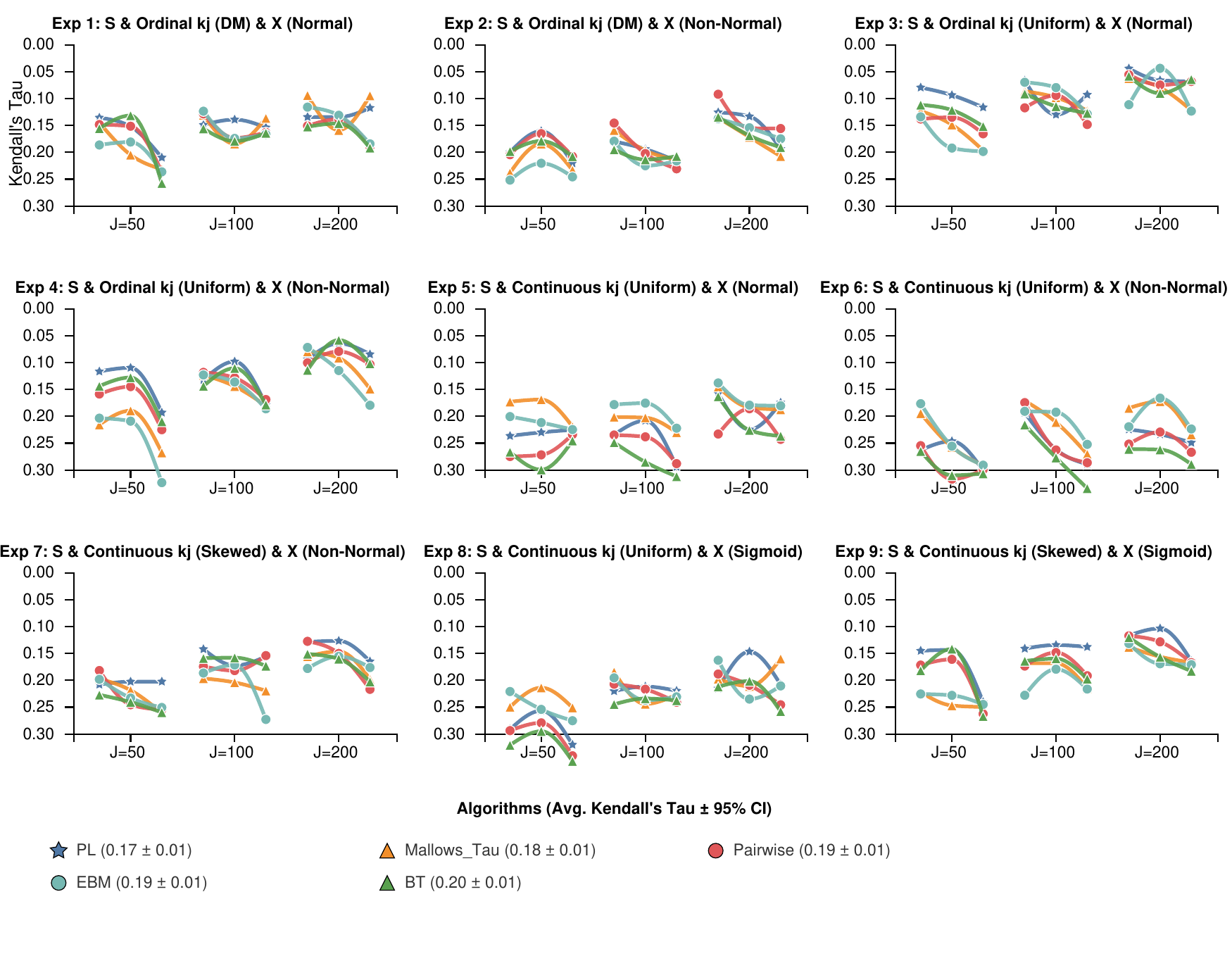} 
    \vspace{-2em}
    \caption{Kendall's tau distance (ordering) for PL generated data
    }
    \label{fig:pl_tau} 
\end{figure*}

\begin{figure*}[htbp]
    \centering \includegraphics[width=1.0\linewidth]{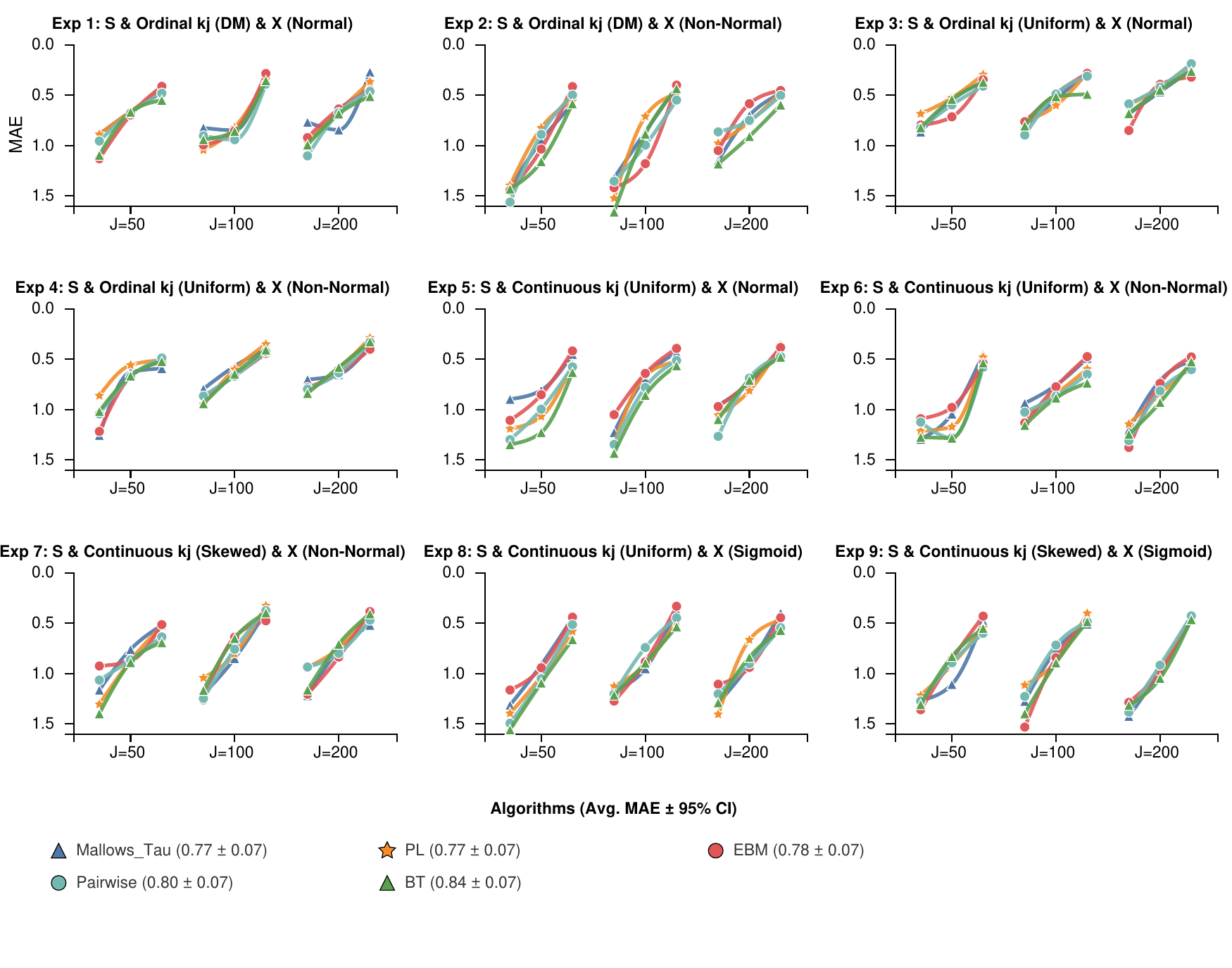} 
    \vspace{-2em}
    \caption{MAE (staging) for PL generated data
    }
    \label{fig:pl_mae} 
\end{figure*}

\begin{figure*}[htbp]
    \centering \includegraphics[width=1.0\linewidth]{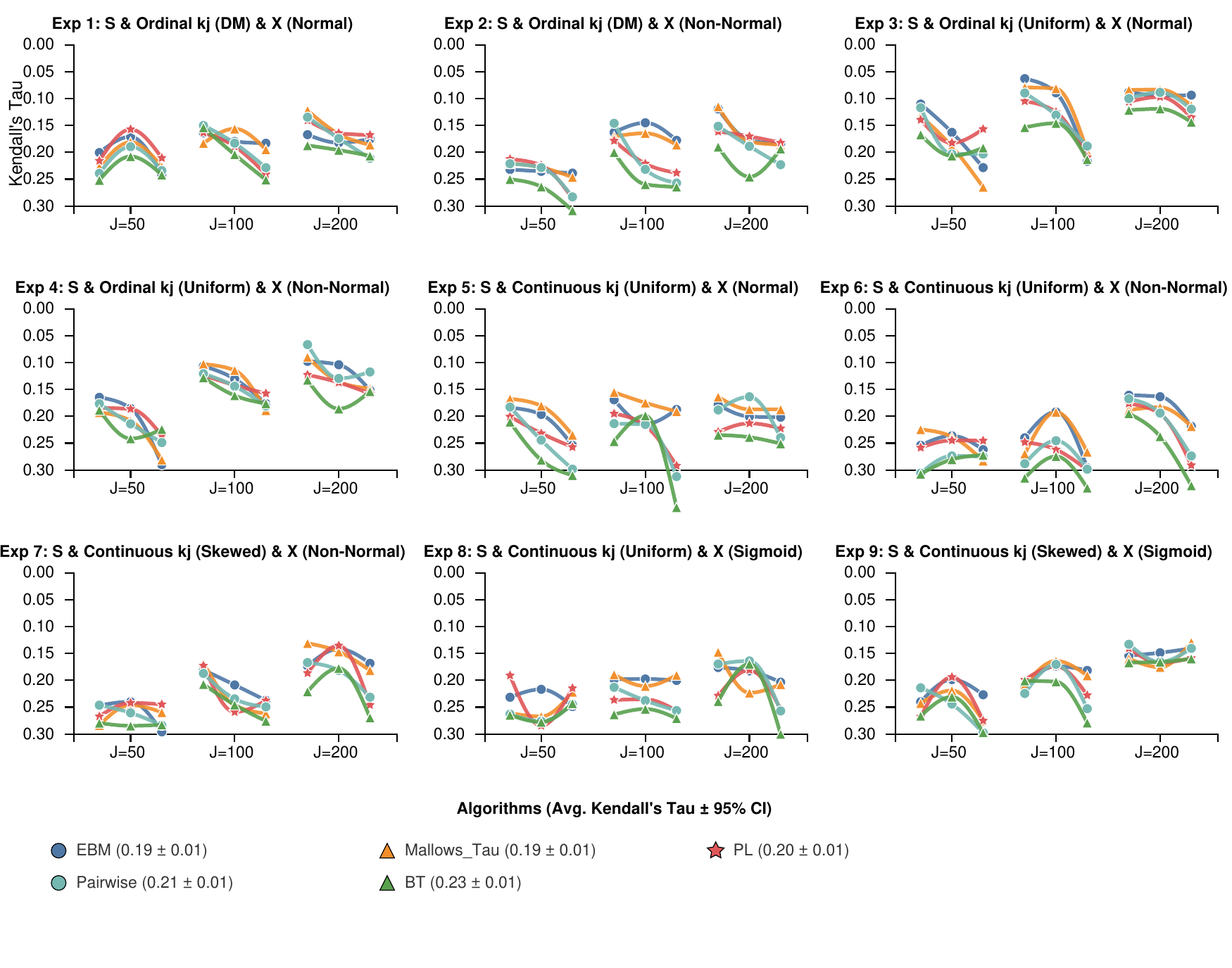} 
    \vspace{-2em}
    \caption{Kendall's tau distance (ordering) for Mallows (dispersion = 1.0) generated data
    }
    \label{fig:m1_tau} 
\end{figure*}

\begin{figure*}[htbp]
    \centering \includegraphics[width=1.0\linewidth]{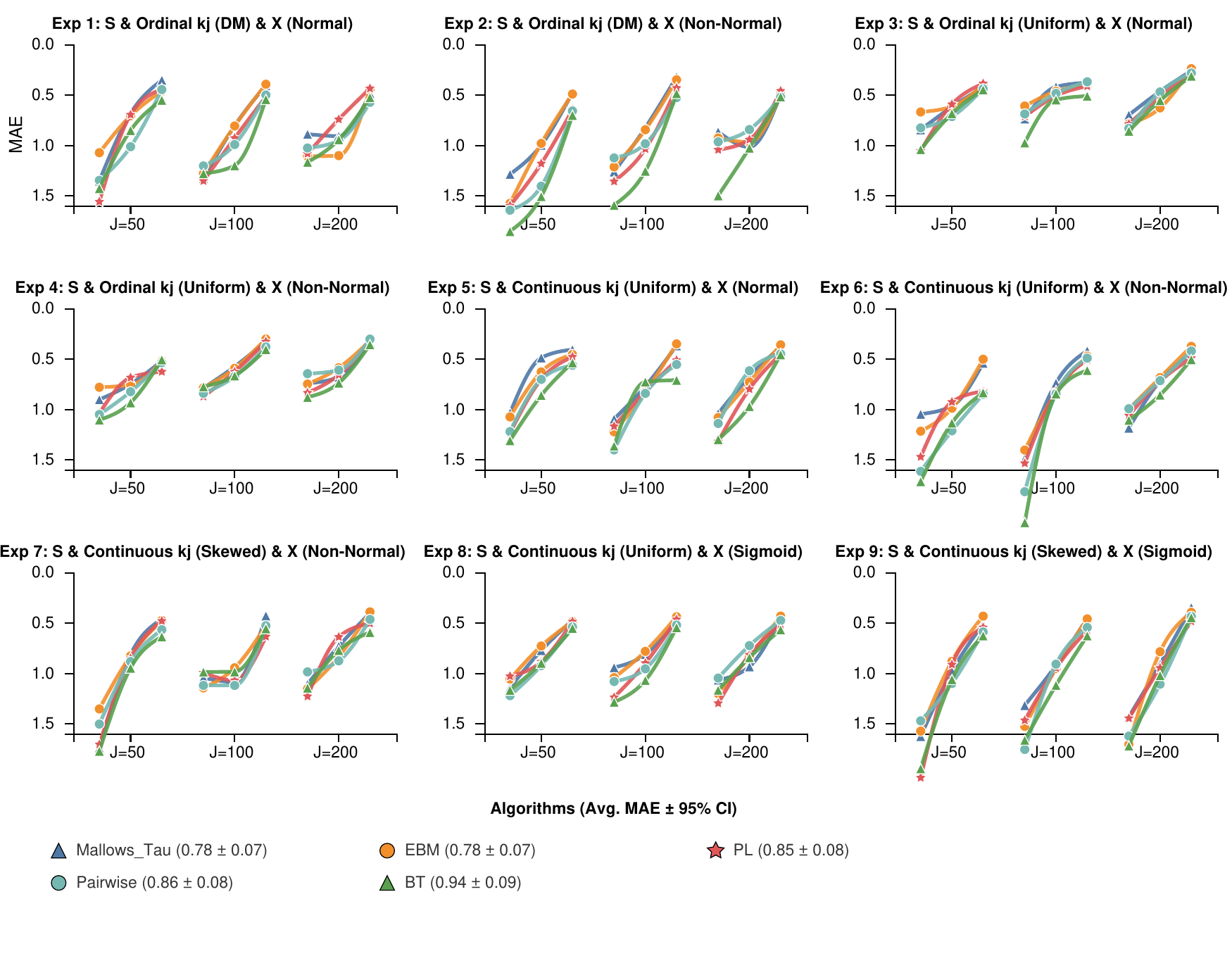} 
    \vspace{-2em}
    \caption{MAE (staging) for Mallows (dispersion = 1.0) generated data
    }
    \label{fig:m1_mae} 
\end{figure*}

\begin{figure*}[htbp]
    \centering \includegraphics[width=1.0\linewidth]{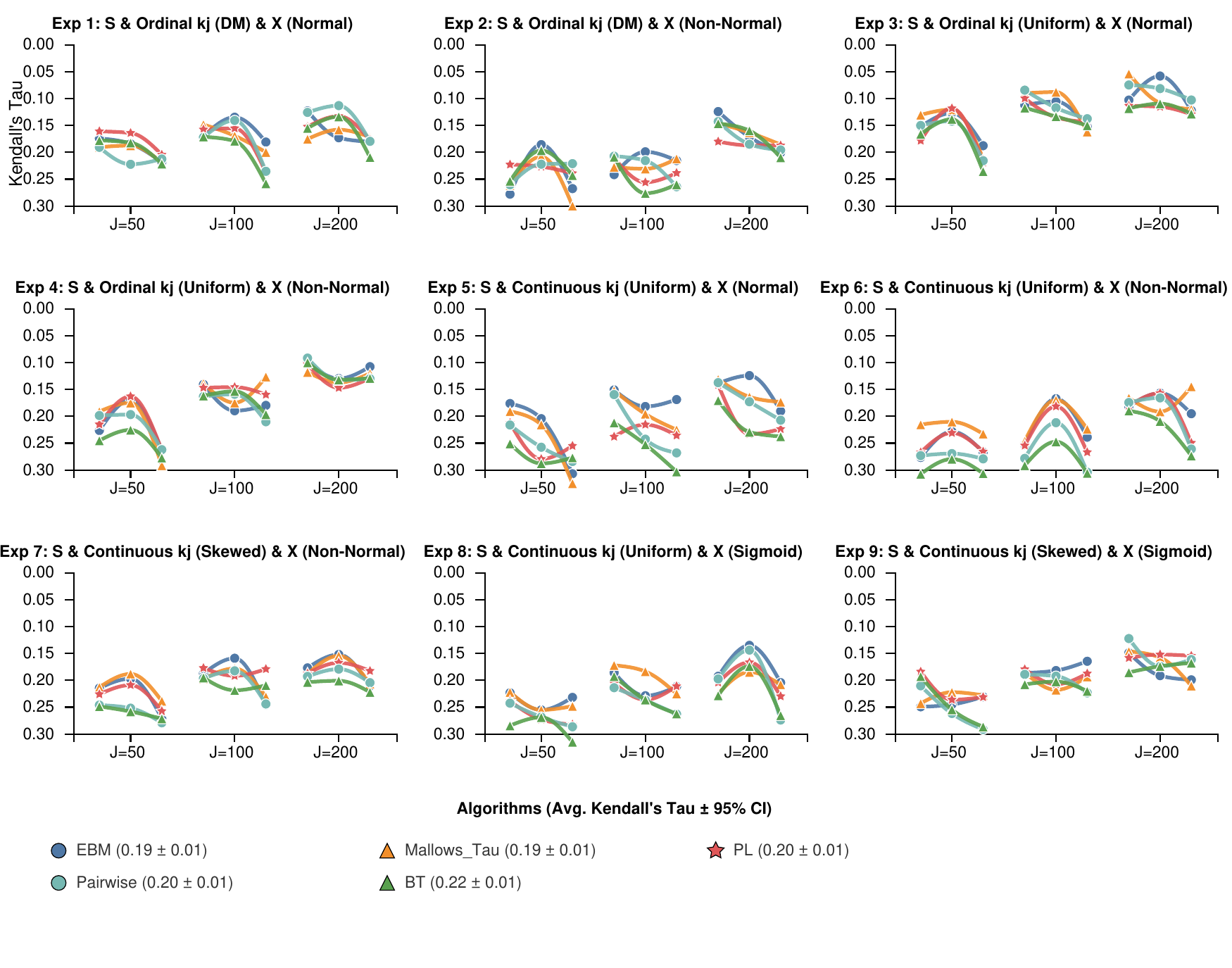} 
    \vspace{-2em}
    \caption{Kendall's tau distance (ordering) for Mallows ($\theta$ = 10.0) generated data
    }
    \label{fig:m10_tau} 
\end{figure*}

\begin{figure*}[htbp]
    \centering \includegraphics[width=1.0\linewidth]{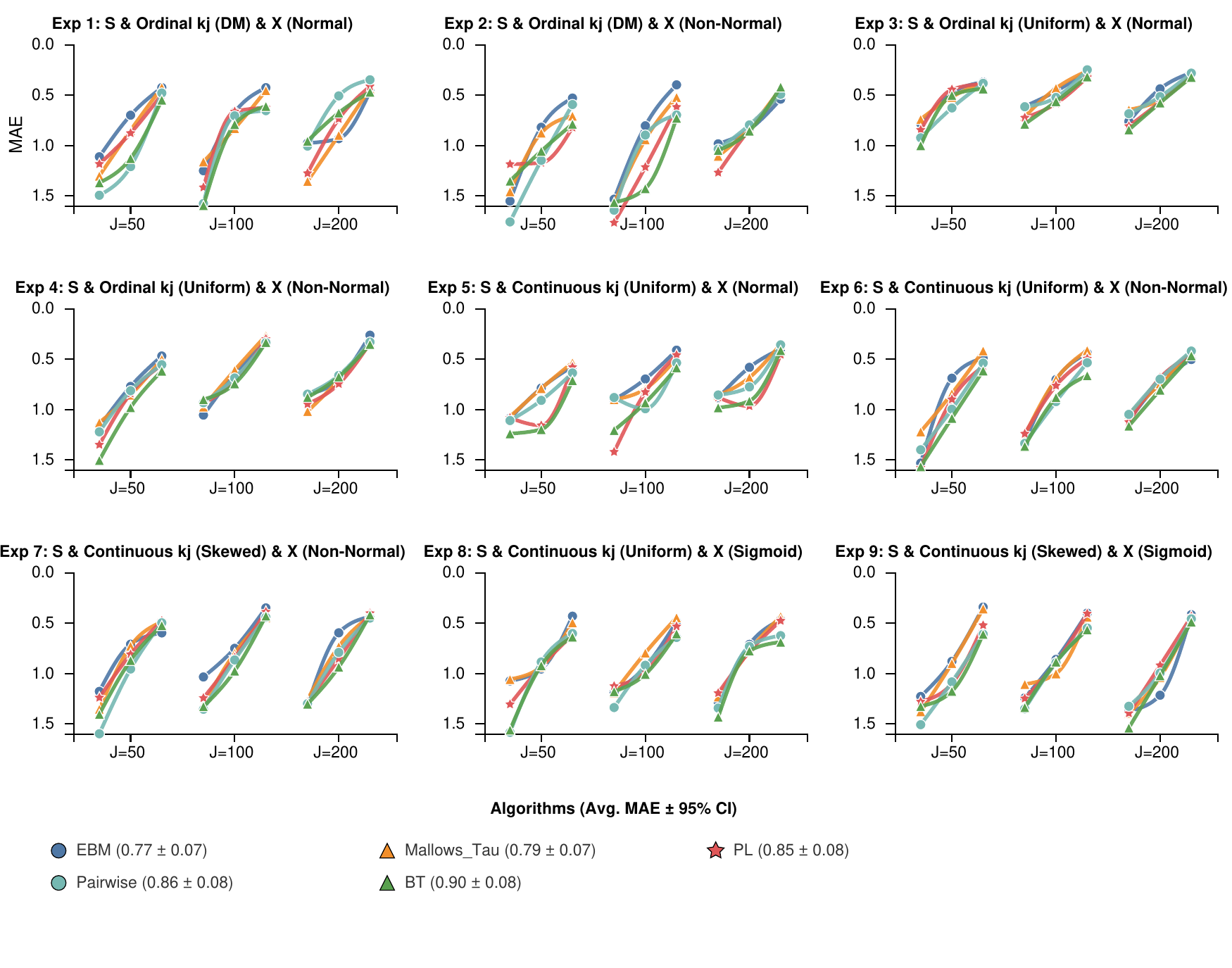} 
    \vspace{-2em}
    \caption{MAE (staging) for Mallows ($\theta$ = 10.0) generated data
    }
    \label{fig:m10_mae} 
\end{figure*}

\clearpage 
\section{NACC Data Preprocessing Pipeline}
\label{apd:nacc_data}

This appendix provides a comprehensive description of the data processing pipeline used to select and prepare the National Alzheimer's Coordinating Center (NACC) dataset for our real-world experiments. The process involved three main stages: cohort definition, biomarker data integration, and final data transformation.

The following data are needed:

\begin{itemize}
  \item UDS with NP and Genetics plus FTLD and LBD modules: \texttt{ftldlbd\_nacc70.csv}
  \item CSF: \texttt{fcsf\_nacc70.csv}
  \item Phenotype Harmonization Consortium (PHC) scores with NACCID and visit number: \texttt{ADSP-PHC\_Cognition\_NACC\_DS\_Freeze1.csv}
  \item Mixed protocol (non-SCAN compliant) MRI: \texttt{investigator\_mri\_nacc70.csv}
\end{itemize}

\subsection{Initial Data Loading and Cohort Definition}
The first stage focused on identifying distinct diagnostic cohorts from the primary NACC dataset.

\paragraph{Data Loading and Subject Uniqueness}
We began with the NACC dataset file \texttt{ftldlbd\_nacc70.csv}, which contained longitudinal data for \textbf{54,631 unique participants}. To create a cross-sectional dataset for our analysis, we filtered these records to retain only the earliest visit for each participant, identified by \texttt{NACCVNUM = 1}.

\paragraph{Pathological Cohort Identification}
We then defined four primary pathological cohorts by creating Boolean flags based on a combination of clinical diagnosis and derived NACC variables. A participant was considered positive for a pathology if they met any of the criteria listed below:

\begin{itemize}
    \item \textbf{Alzheimer's Disease (AD):} \texttt{PROBAD=1}, \texttt{POSSAD=1}, \texttt{NACCALZP=1}, or \texttt{NACCALZD=1}.
    \item \textbf{Lewy Body Dementia (LBD):} \texttt{NACCLBDP=1} or \texttt{NACCLBDE=1}.
    \item \textbf{Frontotemporal Lobar Degeneration (FTLD):} \texttt{NACCFTD=1}, \texttt{FTD=1}, \texttt{FTLDNOS=1}, or \texttt{FTLDMO=1}.
    \item \textbf{Vascular Dementia (VaD):} \texttt{NACCVASC=1}, \texttt{VASC=1}, or \texttt{CVD=1}.
\end{itemize}

\subsection{Biomarker Data Integration and Filtering}
The second stage involved merging and filtering data from three separate biomarker files to enrich the primary dataset.

\paragraph{Cerebrospinal Fluid (CSF) Data}
We loaded CSF biomarker data from \texttt{nacc\_csf.csv}. To ensure data quality and relevance, we applied two key filters. First, we selected only baseline visit records. Second, because CSF sample dates do not always align perfectly with clinical visit dates, we implemented a temporal window, associating a CSF sample with a visit only if the sample was taken within \textbf{180 days} of the visit date. The resulting data was then merged with our main participant dataframe.

\paragraph{Phenotype Harmonization Consortium (PHC) Data}
Again, we selected only baseline visit records. This dataset was then merged directly with the main dataframe using the participant ID as the key.

\paragraph{MRI Volumetric Data}
Structural MRI data was loaded from \texttt{mri\_vol.csv}. We filtered this dataset to include only baseline MRI scans before merging it with the main dataframe on the participant ID.

\subsection{Final Cohort Analysis and Data Availability}
After successfully merging all data sources, we finalized the cohorts and quantified the availability of biomarker data across them.

\paragraph{Creation of Mutually Exclusive Cohorts}
Using the pathological flags defined earlier, we created a set of mutually exclusive cohorts. This allowed us to distinguish between participants with a single pathology (e.g., \texttt{AD only}) and those with mixed pathologies (e.g., \texttt{AD VaD}).

\paragraph{Layered Availability Analysis}
We generated boolean flags for each participant to indicate the presence of valid data for each biomarker type (\texttt{has\_csf}, \texttt{has\_phc}, \texttt{has\_mri}). We then systematically counted the number of participants within each mutually exclusive cohort that had each possible combination of these data types. The results of this analysis are presented in Table~\ref{tab:nacc_availability}, which provides a detailed view of data completeness across the study population.

\subsection{Data Transformation for Modeling}
As a final preprocessing step, the fully integrated and filtered dataframe, which was in a wide format (one column per biomarker), was transformed into a long format using the \texttt{pandas.melt} function. This reshaping is a standard and necessary step to structure the data appropriately for subsequent statistical modeling and analysis.

The final overall data availability after our processing is displayed in Table~\ref{tab:nacc_availability}. The biomarkers chosen for AD and VaD, with detailed information about each biomarker, are available in Table~\ref{tab:nacc_biomarker_glossary}. 

\begin{table*}[ht]
\centering
\caption{Layered Availability of NACC Data Across Cohorts. This table shows the total number of participants in each diagnostic cohort and the count of participants for whom specific data types (MRI, CSF, PHC) are available, both individually and in combination.}
\label{tab:nacc_availability}
\resizebox{\linewidth}{!}{%
\begin{tabular}{lrrrrrrrr}
\toprule
\textbf{Cohort} & \textbf{Total} & \textbf{+MRI} & \textbf{+CSF} & \textbf{+PHC} & \textbf{+MRI+PHC} & \textbf{+CSF+MRI} & \textbf{+CSF+PHC} & \textbf{+CSF+MRI+PHC} \\
\midrule
healthy & 26355 & 4593 & 1094 & 18119 & 3746 & 234 & 990 & 205 \\
AD\_only & 13400 & 1656 & 463 & 10023 & 1474 & 87 & 413 & 82 \\
AD\_VaD & 4449 & 630 & 276 & 3448 & 578 & 53 & 255 & 50 \\
VaD\_only & 3903 & 581 & 246 & 3171 & 532 & 40 & 236 & 36 \\
FTLD\_only & 2379 & 146 & 27 & 1024 & 93 & 6 & 17 & 5 \\
LBD\_only & 1140 & 89 & 30 & 835 & 74 & 5 & 24 & 5 \\
FTLD\_VaD & 817 & 50 & 10 & 505 & 35 & 1 & 4 & 1 \\
AD\_FTLD & 682 & 39 & 19 & 333 & 27 & 4 & 14 & 3 \\
AD\_LBD & 516 & 48 & 12 & 358 & 43 & 4 & 11 & 3 \\
LBD\_VaD & 422 & 64 & 20 & 341 & 58 & 2 & 16 & 2 \\
AD\_LBD\_VaD & 278 & 27 & 10 & 211 & 24 & 1 & 10 & 1 \\
AD\_FTLD\_VaD & 227 & 12 & 8 & 150 & 11 & 0 & 5 & 0 \\
FTLD\_LBD & 26 & 0 & 0 & 12 & 0 & 0 & 0 & 0 \\
AD\_FTLD\_LBD & 15 & 0 & 0 & 7 & 0 & 0 & 0 & 0 \\
FTLD\_LBD\_VaD & 13 & 0 & 0 & 6 & 0 & 0 & 0 & 0 \\
AD\_FTLD\_LBD\_VaD & 9 & 0 & 0 & 5 & 0 & 0 & 0 & 0 \\
\bottomrule
\end{tabular}%
}
\end{table*}

\begin{table*}[ht]
\centering
\caption{Glossary of NACC Biomarkers with Source, Units, and Interpretation}
\label{tab:nacc_biomarker_glossary}
\resizebox{\linewidth}{!}{%
\begin{tabular}{lllllll}
\hline
\textbf{Abbrev.} & \textbf{Full Name} & \textbf{Source} & \textbf{Unit / Scale} & \textbf{Higher Values Indicate} & \textbf{AD} & \textbf{VaD} \\
\hline
ptau\_abeta     & Phospho-Tau$_{181}$/A$\beta_{1-42}$ ratio & CSF & Ratio & More pathology (higher tau-to-amyloid) & True & False \\
ttau\_abeta     & Total Tau/A$\beta_{1-42}$ ratio           & CSF & Ratio & More pathology (higher tau-to-amyloid) & True & False \\
CSFABETA        & Amyloid Beta (A$\beta_{1-42}$)            & CSF & pg/mL & Less pathology (lower deposition) & True & False \\
CSFPTAU         & Phosphorylated Tau$_{181}$                & CSF & pg/mL & More pathology (tangle burden) & True & False \\
CSFTTAU         & Total Tau                                 & CSF & pg/mL & More pathology (neurodegeneration) & True & False \\
HippVolNorm     & Hippocampal Volume (normalized)            & MRI & Ratio & Less pathology (larger volume) & True & False \\
EntorhinalNorm  & Entorhinal Volume (normalized)            & MRI & Ratio & Less pathology (larger volume) & True & False \\
MidTempNorm     & Middle Temporal Volume (normalized)       & MRI & Ratio & Less pathology (larger volume) & True & False \\
FusiformNorm    & Fusiform Volume (normalized)              & MRI & Ratio & Less pathology (larger volume) & True & False \\
WholeBrainNorm  & Whole Brain Volume (normalized)           & MRI & Ratio & Less pathology (larger volume) & True & True \\
VentricleNorm   & Ventricular Volume (normalized)           & MRI & Ratio & More pathology (enlargement) & True & True \\
WMHnorm         & White Matter Hyperintensity (normalized)  & MRI & Ratio & More pathology (greater burden) & False & True \\
PHC\_MEM        & Harmonized Composite Memory Score         & Cognitive composite & z-score & Better cognition & True & True \\
PHC\_EXF        & Harmonized Composite Executive Function   & Cognitive composite & z-score & Better cognition & True & True \\
PHC\_LAN        & Harmonized Composite Language Score       & Cognitive composite & z-score & Better cognition & True & True \\
\hline
\end{tabular}
}
\end{table*}

We included the two ratios, i.e., \texttt{ptau\_abeta} and \texttt{ttau\_abeta} because \cite{palmqvist2015detailed} found that these ratios have higher diagnostic power for Alzheimer's disease than raw Abeta values. 

\clearpage
\section{NACC results}
\label{apd:nacc_results}

\begin{figure}[htbp]
    \centering 
    \includegraphics[width=1.0\linewidth]{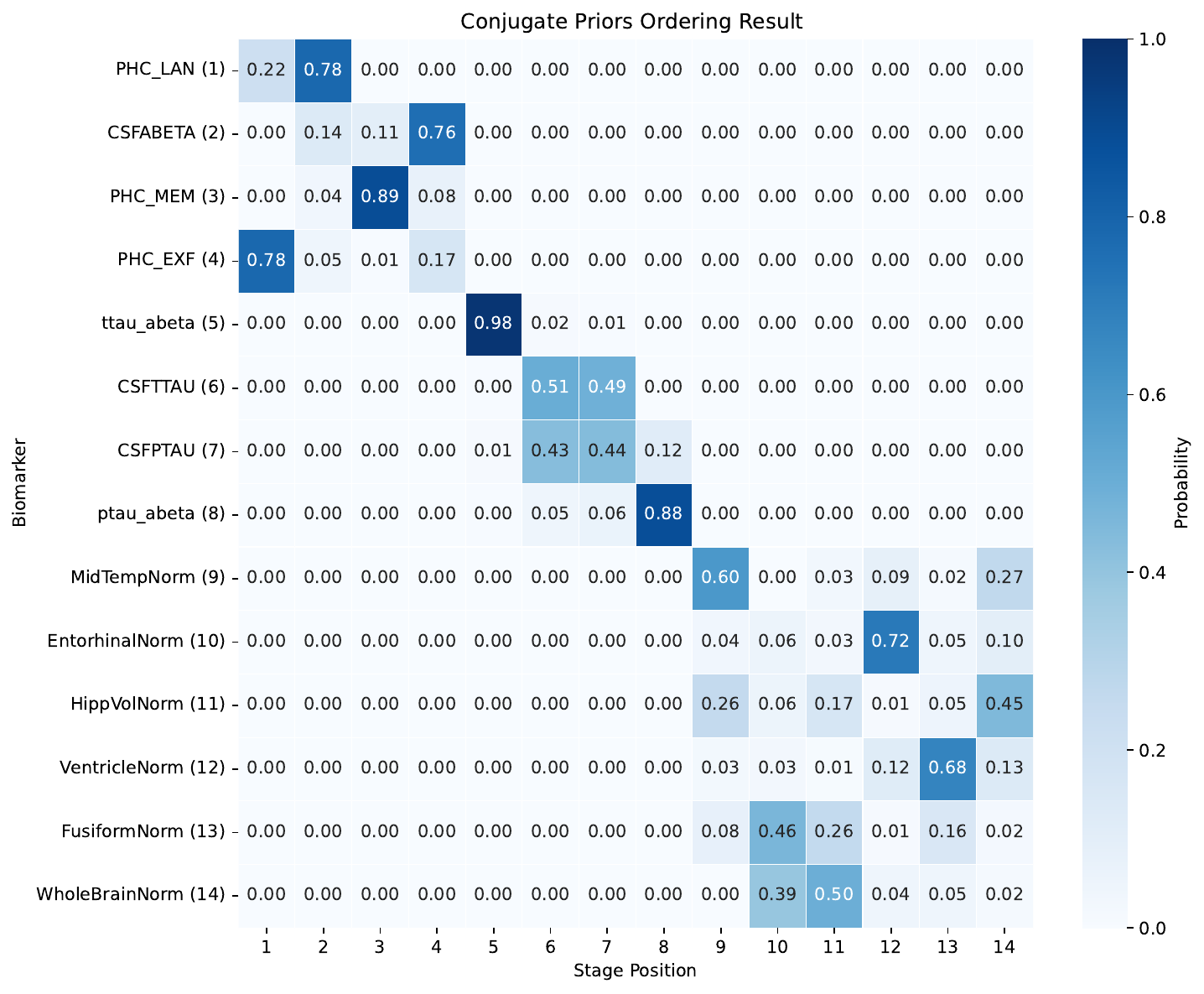} 
    \vspace{-2em}
    \caption{Progression of AD, obtained via SA-EBM
    }
    \label{fig:ad_progression} 
\end{figure}

\begin{figure}[htbp]
    \centering 
    \includegraphics[width=0.9\linewidth]{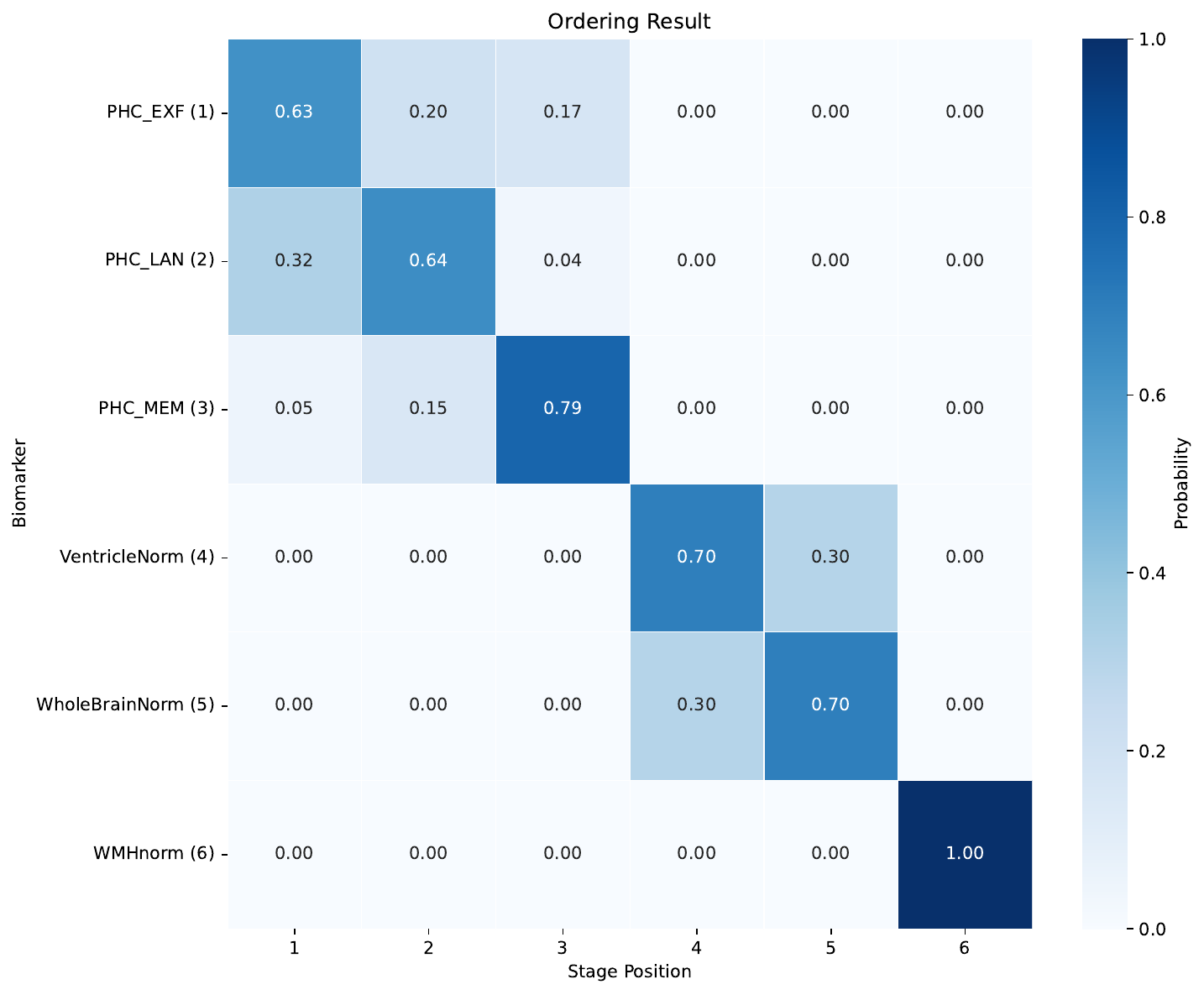} 
    \caption{Progression of VaD, obtained via SA-EBM
    }
    \label{fig:vad_progression} 
\end{figure}

\begin{figure}[htbp]
    \centering 
    \includegraphics[width=1.0\linewidth]{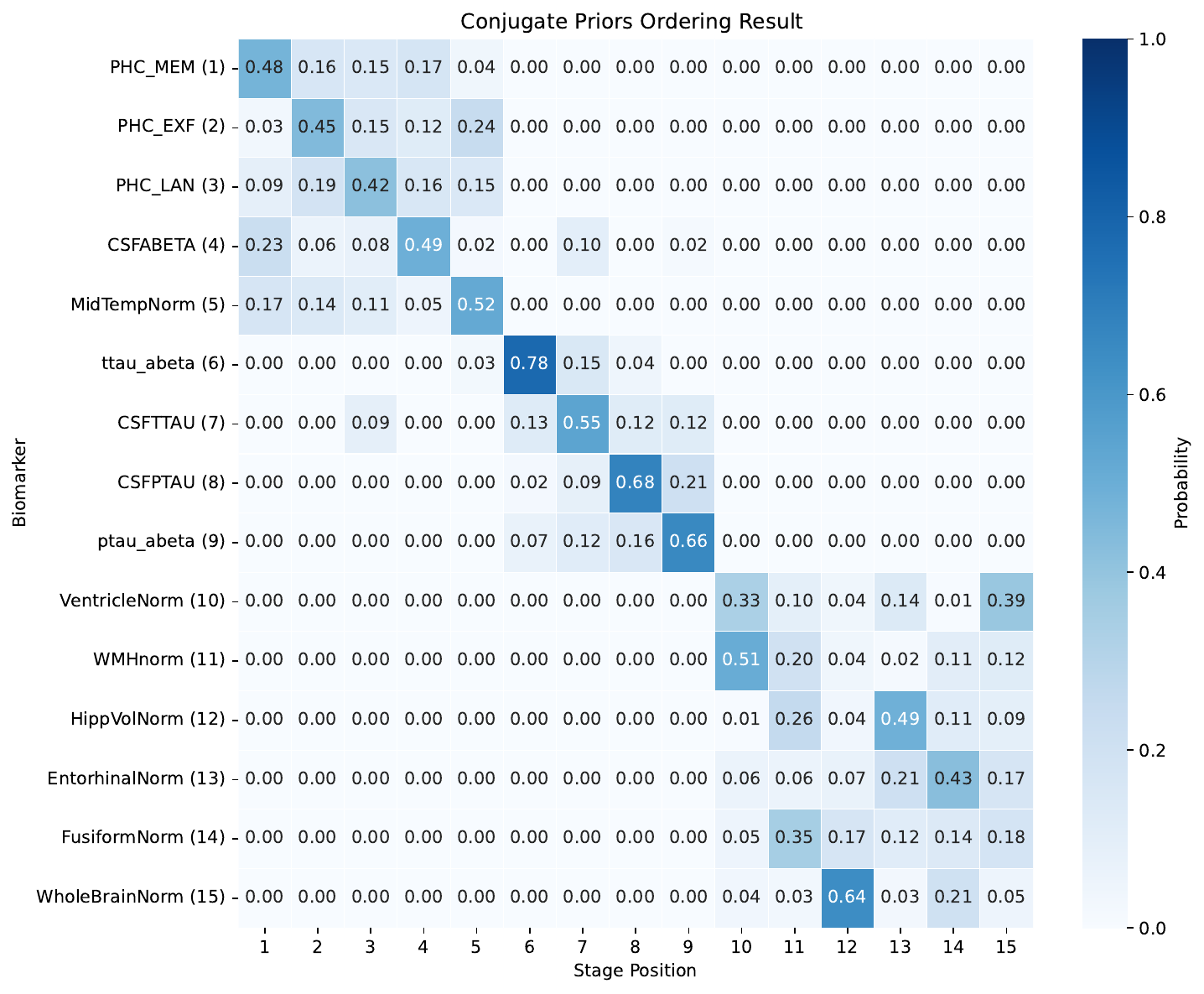} 
    \vspace{-2em}
    \caption{Progression of the mixed pathology of AD and VaD, obtained via SA-EBM
    }
    \label{fig:mp_progression} 
\end{figure}

\begin{figure}[htbp]
    \centering 
    \includegraphics[width=1.0\linewidth]{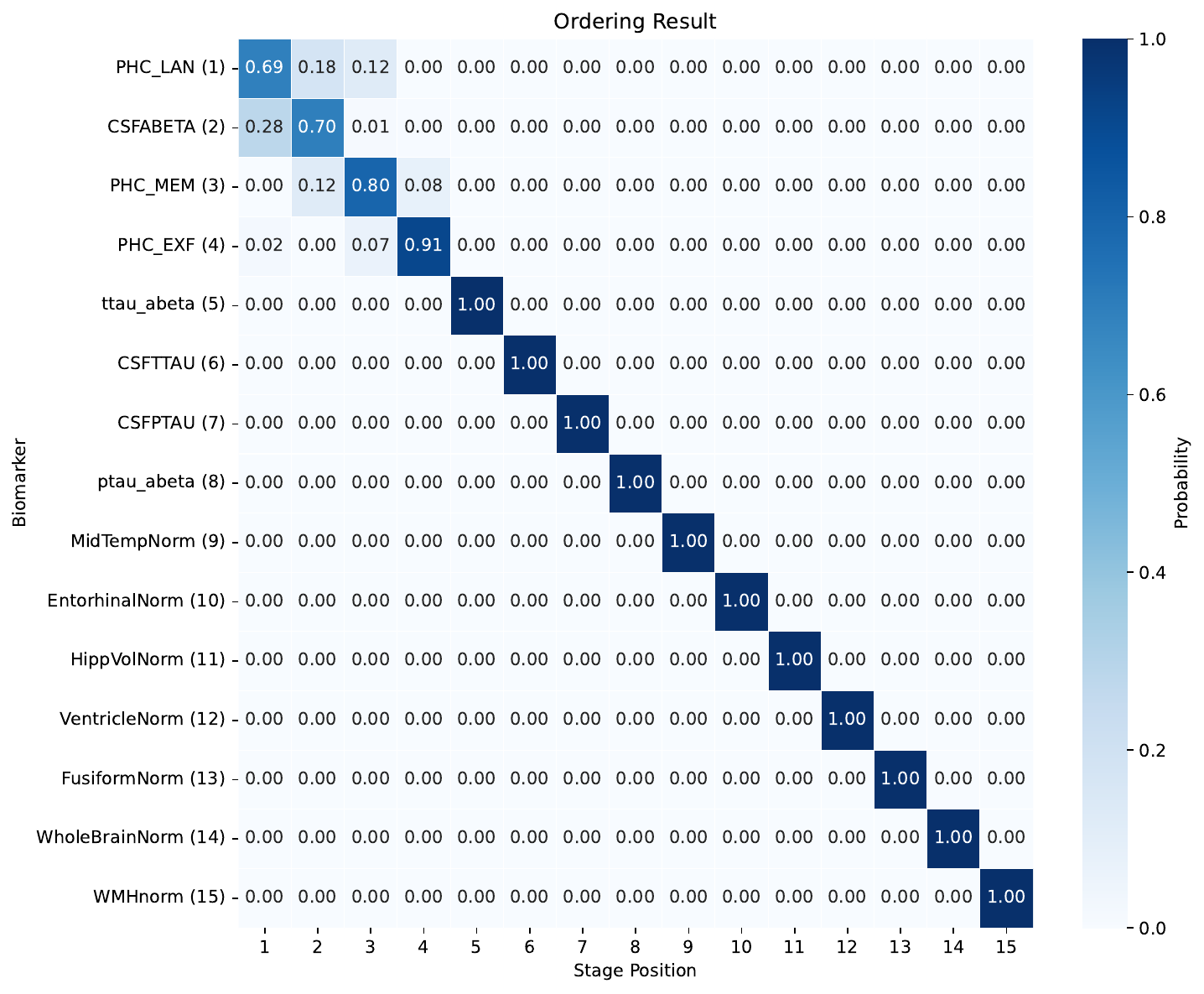} 
    \vspace{-2em}
    \caption{Progression of the mixed pathology of AD and VaD, obtained via JPM (BT)
    }
    \label{fig:mp_progression_bt} 
\end{figure}

\begin{figure}[htbp]
    \centering 
    \includegraphics[width=1.0\linewidth]{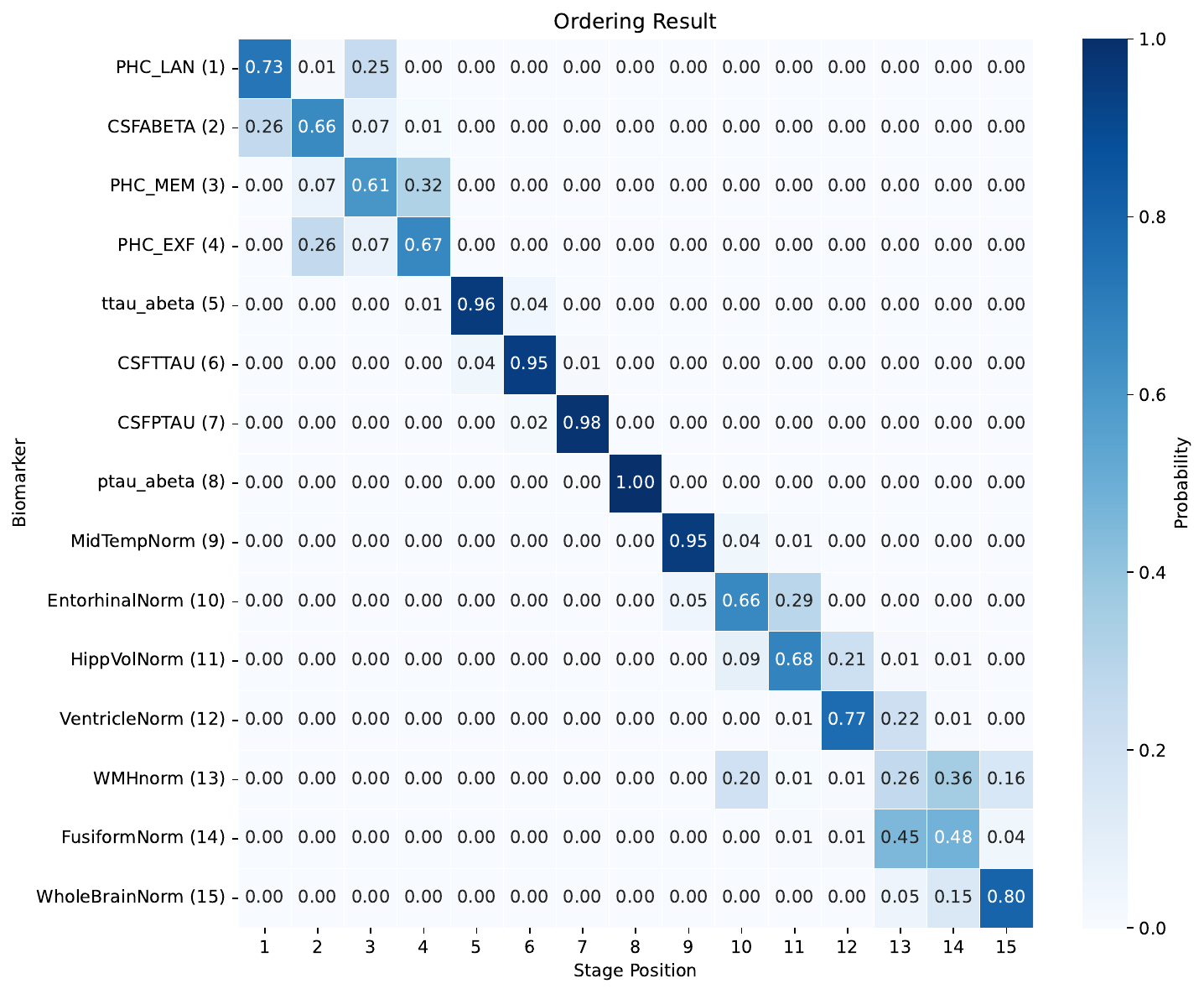} 
    \vspace{-2em}
    \caption{Progression of the mixed pathology of AD and VaD, obtained via JPM (PP)
    }
    \label{fig:mp_progression_pp} 
\end{figure}

\begin{figure}[htbp]
    \centering 
    \includegraphics[width=1.0\linewidth]{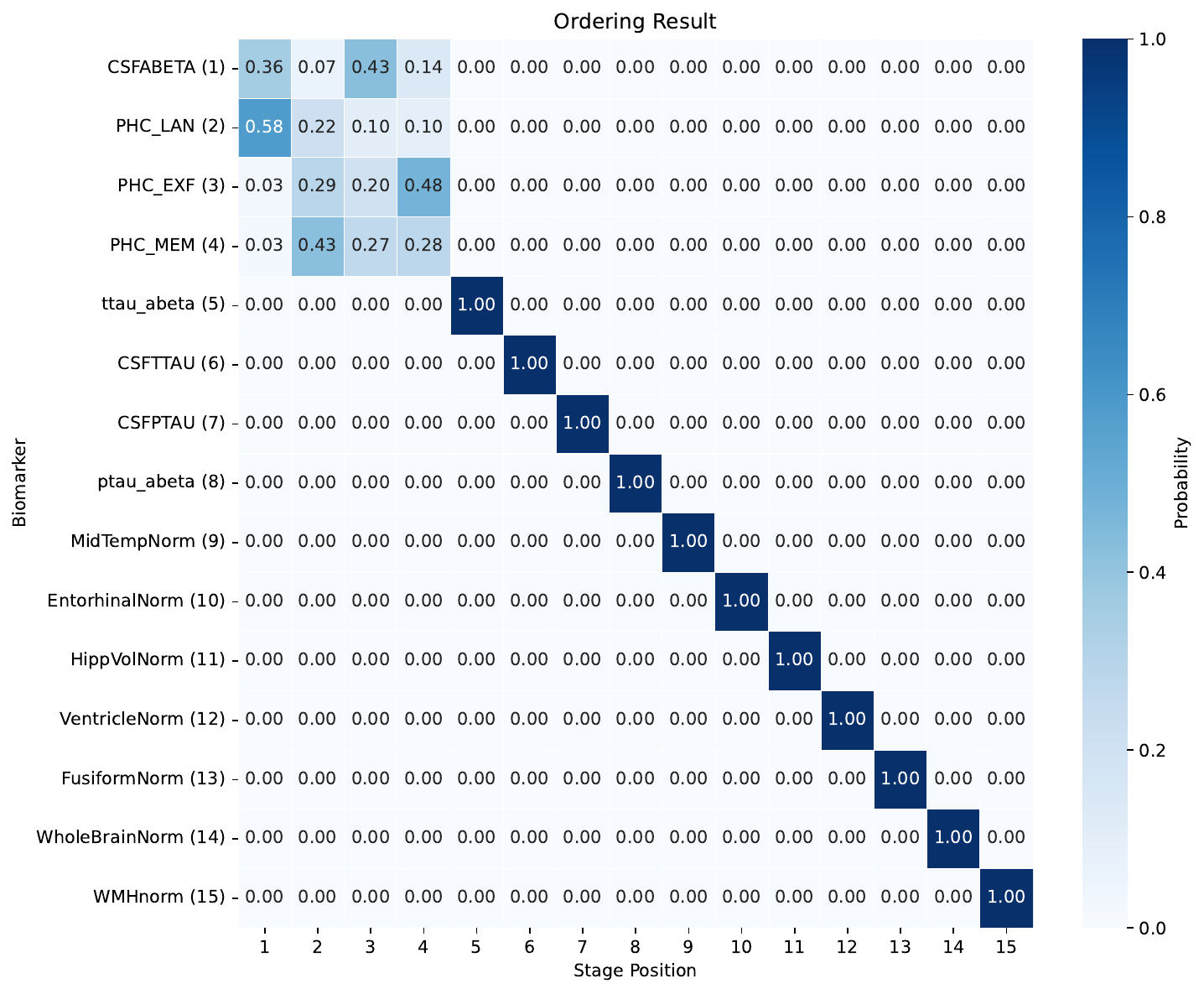} 
    \vspace{-2em}
    \caption{Progression of the mixed pathology of AD and VaD, obtained via JPM (PL)
    }
    \label{fig:mp_progression_pl} 
\end{figure}

\begin{figure}[htbp]
    \centering 
    \includegraphics[width=1.0\linewidth]{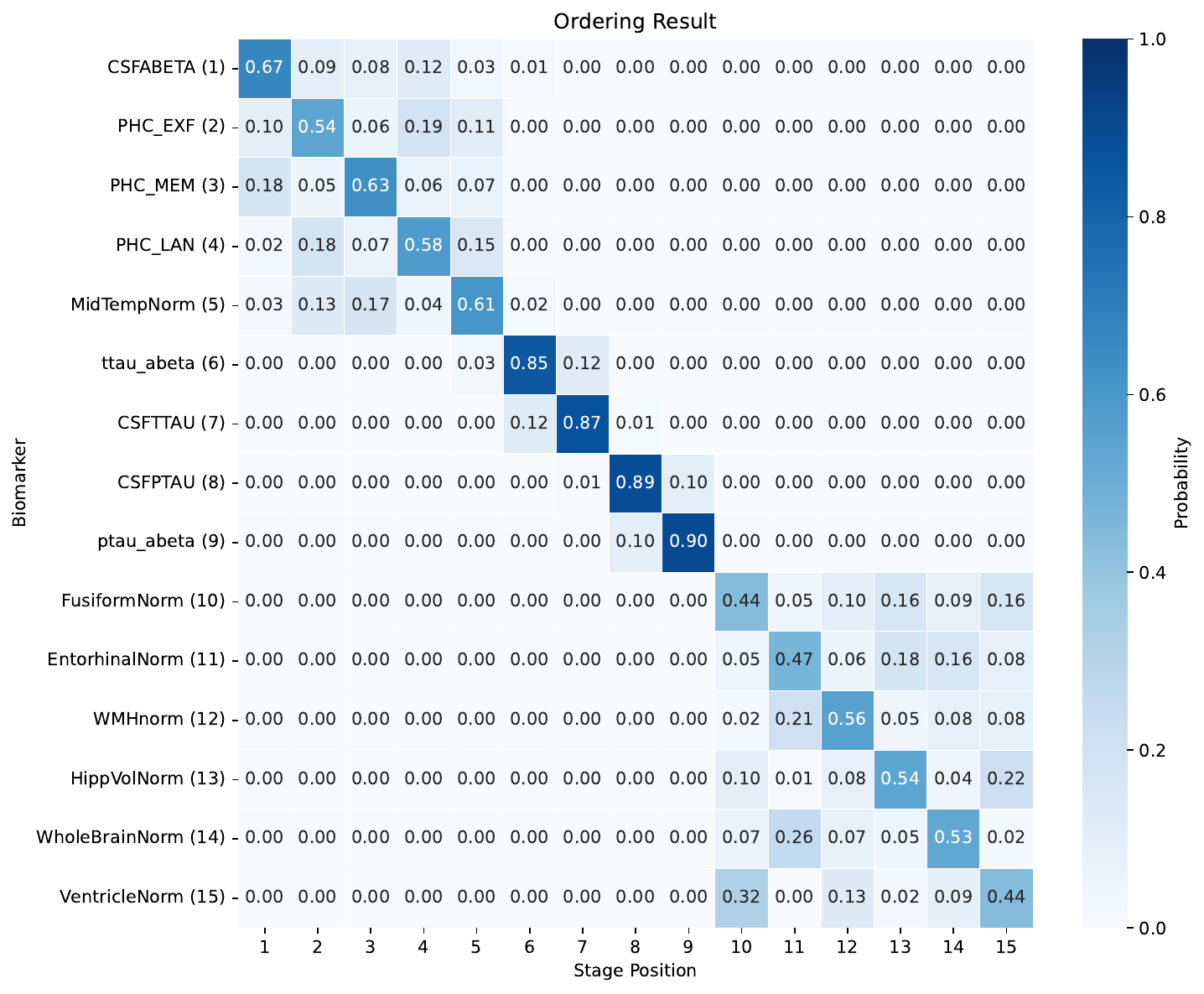} 
    \vspace{-2em}
    \caption{Progression of the mixed pathology of AD and VaD, obtained via JPM (Mallows)
    }
    \label{fig:mp_progression_mallows} 
\end{figure}

\begin{figure}[htbp]
    \centering 
    \includegraphics[width=1.0\linewidth]{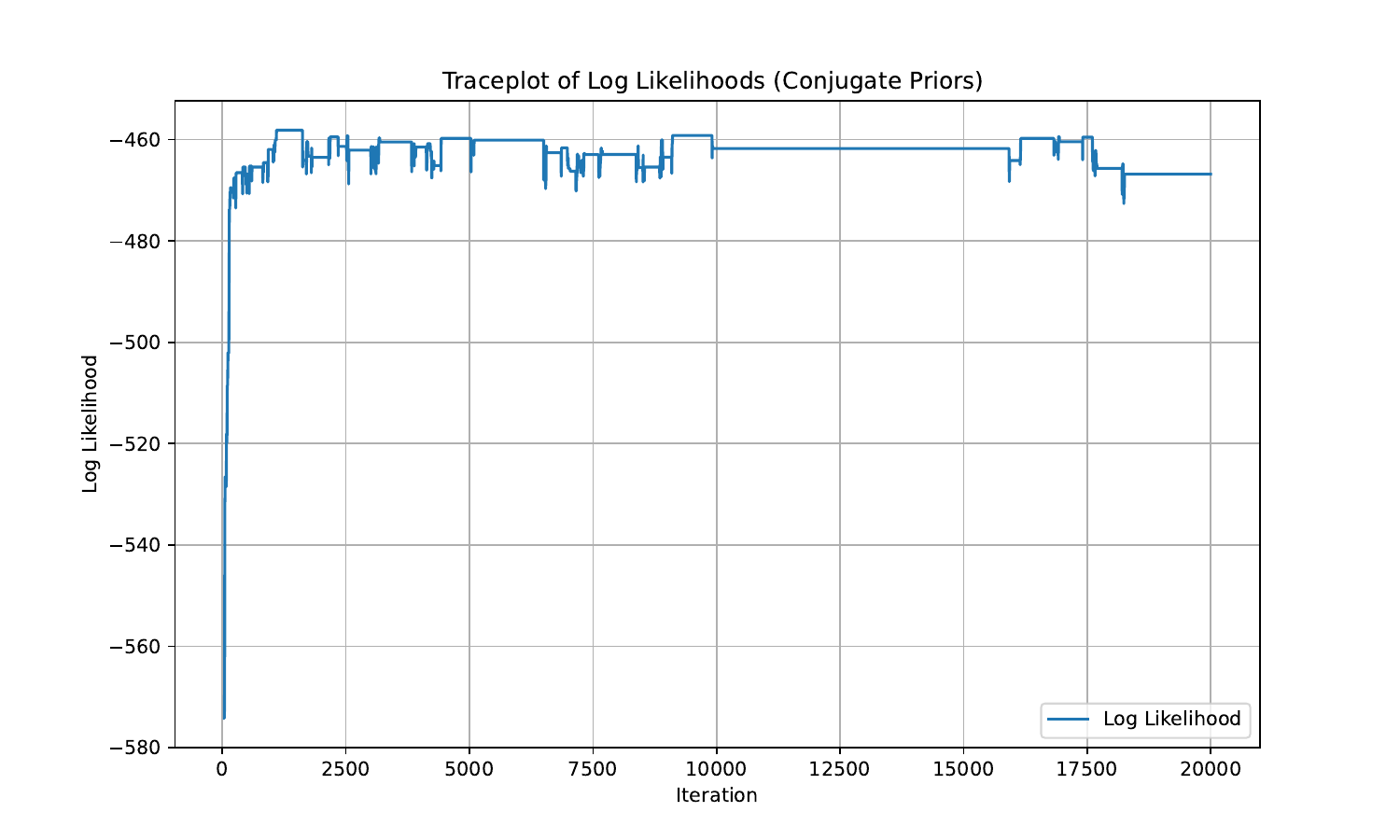} 
    \vspace{-2em}
    \caption{Trace plot of running SA-EBM on the mixed-pathology data
    }
    \label{fig:traceplot_saebm} 
\end{figure}

\begin{figure}[htbp]
    \centering 
    \includegraphics[width=1.0\linewidth]{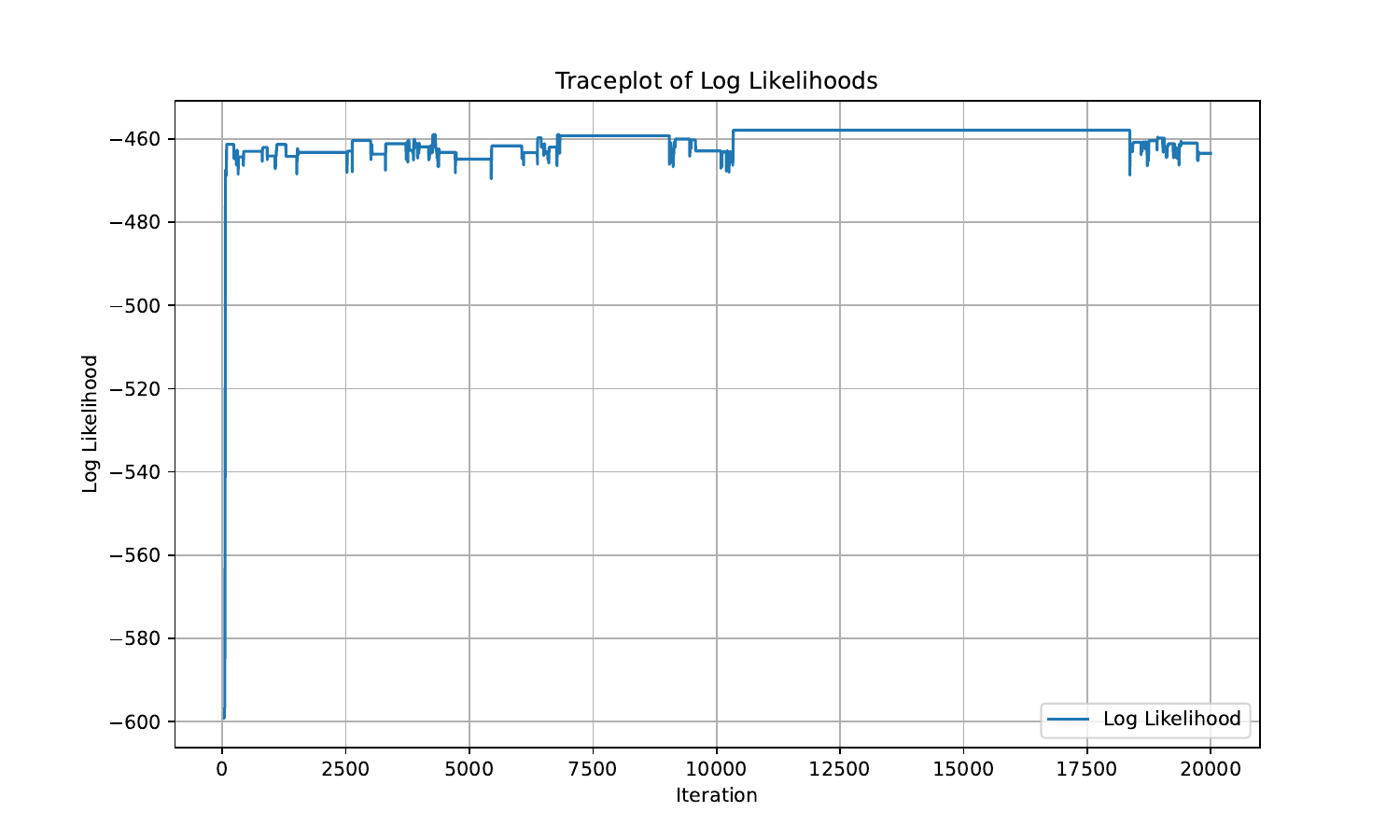} 
    \vspace{-2em}
    \caption{Trace plot of running JPM-Mallows on the mixed-pathology data
    }
    \label{fig:traceplot_mallows} 
\end{figure}

\end{document}